%% file: acl2021.tex
\documentclass[11pt,a4paper]{article}
\usepackage[hyperref]{acl2021}
\usepackage{times}
\usepackage{latexsym}

\usepackage{xspace}
\usepackage[ruled,noline,linesnumbered]{algorithm2e}
\usepackage{algorithm2e}
\usepackage{algorithmic}
\usepackage{microtype}
\usepackage{xcolor}
\usepackage{colortbl}
\usepackage{hhline}
\usepackage{amsmath}
\usepackage{amssymb}
\usepackage{multirow}
\usepackage{booktabs}
\usepackage{graphicx}
\usepackage{enumitem}
\usepackage{tikz}
\usepackage{booktabs}
\usepackage{wrapfig}
\usepackage{fancyhdr}

\newcommand{\method}{\textsc{Reflective Decoding}\xspace}
\newcommand{\rdexp}{\textsc{Reflective Decoder}\xspace}

\newcommand{\decoder}{\method sampling function\xspace}

\newcommand{\lrLM}{$\protect\overrightarrow{\text{LM}}$\xspace}
\newcommand{\rlLM}{$\protect\overleftarrow{\text{LM}}$\xspace}

\newcommand{\rlRD}{$\protect\overleftarrow{\text{RD}}$\xspace}
\newcommand{\lrRD}{$\protect\overrightarrow{\text{RD}}$\xspace}

\newcommand{\TERP}{$\text{TER}_P$\xspace}

\newcommand{\stepone}{contextualization step\xspace}

\newcommand{\steptwo}{reflection step\xspace}

\newcommand{\peter}[1]{\textcolor{blue}{}}
\newcommand{\ari}[1]{\textcolor{purple}{}}

\aclfinalcopy 
\def\aclpaperid{2791} 

\title{Reflective Decoding: Beyond Unidirectional Generation with \\ Off-the-Shelf Language Models}

\author{Peter West\textsuperscript{1,2} \hspace{.1cm} Ximing Lu\textsuperscript{1,2} \hspace{.1cm} Ari Holtzman\textsuperscript{1}\\ 
\textbf{Chandra Bhagavatula\textsuperscript{2}  \hspace{.1cm} Jena Hwang\textsuperscript{2}  \hspace{.1cm} Yejin Choi\textsuperscript{1,2}}\\
  \textsuperscript{1}Paul G. Allen School of Computer Science \& Engineering, University of Washington\\
  \textsuperscript{2}Allen Institute for Artificial Intelligence\\
  \texttt{\{pawest, ahai, yejin\}@cs.washington.edu} \\
  \texttt{\{ximinglu, chandrab, jenah\}@allenai.org}
  }

\date{}

\begin{document}
\maketitle

\input{sections/abstract}

\input{sections/introduction}

\input{sections/method}

\input{sections/experiments}

\input{sections/results}

\input{sections/discussion}

\input{sections/background}

\input{sections/conclusion}


\input{sections/acknowledgements}

\input{sections/broader_impact}

\bibliographystyle{acl_natbib}
\bibliography{anthology,acl2021}

\clearpage

\appendix

\input{sections/appendix}

\end{document}

%% file: sections/abstract.tex
\begin{abstract}
Publicly available, 
large pretrained Language Models (LMs) generate text with remarkable quality, but only sequentially from left to right. As a result, they are not immediately applicable to generation tasks that break the unidirectional assumption, such as paraphrasing or text-infilling, necessitating task-specific supervision.


In this paper, we present \method, a novel unsupervised algorithm that allows for direct application of unidirectional LMs to non-sequential tasks. 
Our 2-step approach requires no supervision or even parallel corpora, only two off-the-shelf pretrained LMs in opposite directions: \emph{forward} and \emph{backward}. 
First, in the \emph{contextualization} step, 
we use LMs to generate ensembles of past and future contexts which collectively capture the input (e.g. the source sentence for paraphrasing
). Second, in the \emph{reflection} step, we condition on these ``context ensembles'', generating outputs that are compatible with them. 
Comprehensive empirical results$^{1}$ demonstrate that \method outperforms strong unsupervised baselines on both paraphrasing and abductive text infilling, significantly narrowing the gap between unsupervised and supervised methods. \method surpasses multiple supervised baselines on various metrics including human evaluation.
\end{abstract}

%% file: sections/introduction.tex
\section{Introduction}
\label{introduction}

\footnotetext[1]{Further results and resource are available at \url{https://homes.cs.washington.edu/~pawest/ReflectiveDecoding.html}
}
\addtocounter{footnote}{1}



Language Models (LMs) like GPT-2 \citep{radford2019gpt2}, trained over vast unstructured data, can leverage enhanced generation methods \citep{Holtzman2019TheCC, martins2020sparse, welleck2019neural} to give fluent and coherent continuations to given input text—e.g. news articles or stories. GPT-3 \citep{Brown2020LanguageMA} takes this a step further: given a small number of examples and a well-constructed prompt, it shows remarkable performance on tasks where vast quantities of supervised data and finetuning were thought to be necessary. While this demonstrates the potential for LM-decoding in  few-shot or even zero-shot out-of-the-box settings, limited access to GPT-3 and immense computational cost keep this from being a widely usable or efficient solution. 
\input{figs/fig_1}

Yet recent work shows that GPT-2 may hold similar capabilities when it is primed correctly. \citet{Li2021PrefixTuningOC} achieve supervised-level performance in a few-shot setting using smaller, accessible models like GPT-2. They learn a small number of task-specific vectors as a prefix to the input, without tuning the model itself. Off-the-shelf GPT-2 is capable of few-shot learning given the right setup; our work aims to push this concept further, by showing that out-of-the-box LMs can solve complex generation problems simply by using the right decoding algorithm.

We introduce \textsc{Reflective Decoding}—a novel decoding method that allows LMs to be applied to generation tasks that break the ``text continuation'' paradigm, such as paraphrasing and text-infilling. \method{} requires \textit{no} supervision, only two complementary off-the-shelf LMs—one forward (\lrLM) and one backward (\rlLM). That means no per-task finetuning, even on unstructured text in the target domain.

Inspired by the distributional hypothesis 
\citep{firth1957synopsis}, \method works by generating text that might occupy the \emph{same contexts} as an input.
We use two LMs (\lrLM and \rlLM) to \textit{generate} contexts for a given input, which implicitly capture aspects of its meaning (the \stepone). Then, in the \steptwo, we condition on this ensemble of contexts, decoding \textit{over} the input with generations that are distributionally related to—or replace—the input.

Paraphrasing is a natural application: a good paraphrase should intuitively be compatible with the same contexts as the original text. \method shows strong unsupervised paraphrasing performance: On the Quora question pair dataset, we find one variant of our model ($\text{RD}_{30}$) outperforms unsupervised baselines on all but one metric, and supervised baselines on both the SARI metric and human evaluation. We see the same trends on the Twitter URL corpus \citep{lan2017continuously}. 

\method can also be applied to tasks that only replace part of the input, or generate within it, like \textbf{infilling}; on $\alpha$NLG \citep{Bhagavatula2019AbductiveCR}, we outperform the best unsupervised baseline on overall quality, halving the gap with supervised methods. In both applications, \method directly applies off-the-shelf pretrained models, without finetuning on the task or target domain.
This provides evidence that off-the-shelf Language Models can excel at surprising applications, when paired with decoding algorithms designed to elicit specific kinds of information.



%% file: figs/fig_1.tex
\begin{figure}[t]
    \centering
    \includegraphics[width=\linewidth]{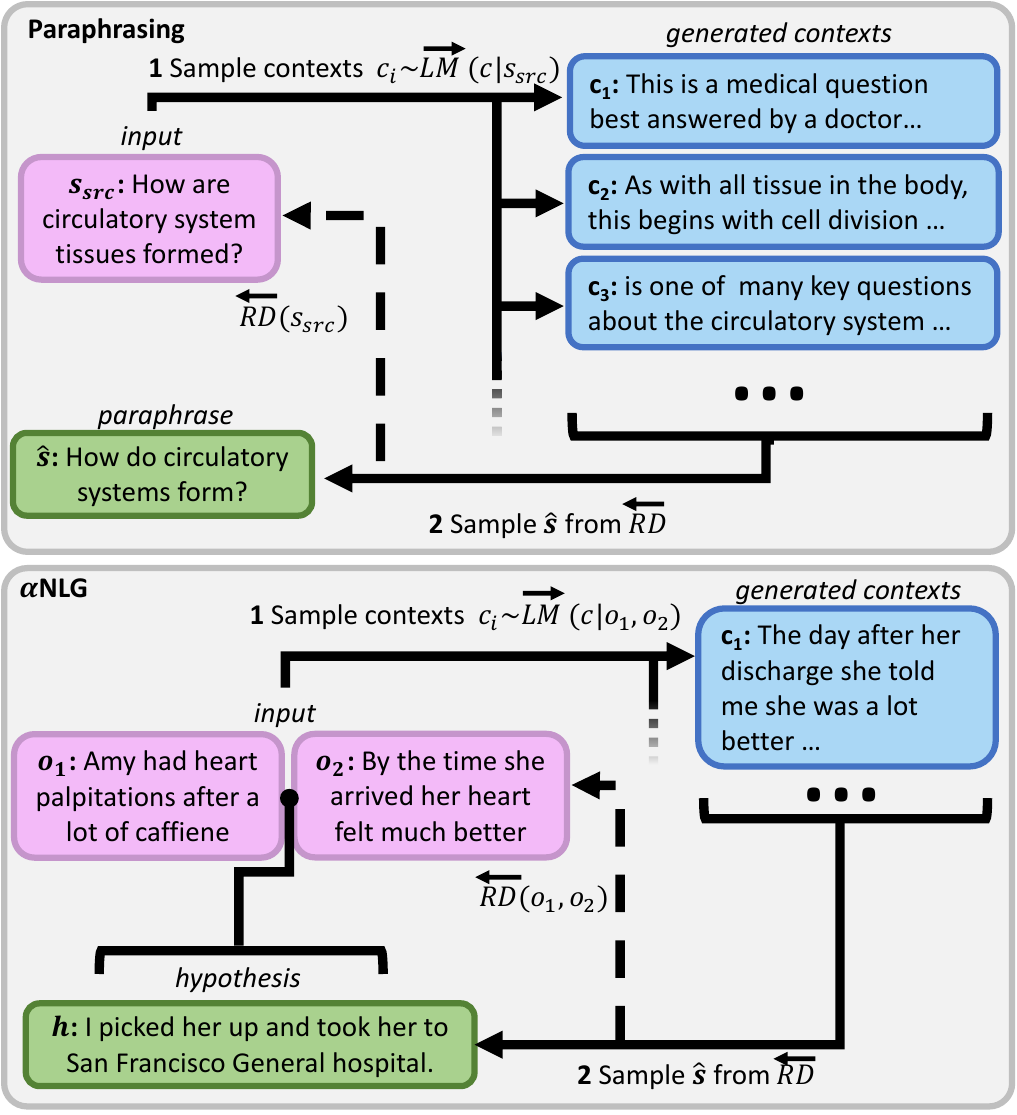}
    \caption{Illustration of \method applied to paraphrasing and abductive infilling ($\alpha$NLG  \citealp{Bhagavatula2019AbductiveCR}). \textbf{Only} the right-context is shown, although \textbf{both} right- and left-contexts are used in practice. First the \stepone (1) captures aspects of an input by generating many representative contexts for it. Then in the \steptwo (2) we sample generations that can replace the input and fit these representative contexts \rlRD. 
    }
    \label{fig_1}
\end{figure}

%% file: sections/method.tex
\section{Method}
\label{sec_method}

\subsection{Notation}
\label{subsec_notation}

Arrows indicate the order in which sampling functions condition on and generate tokens: $\xrightarrow{}$ indicates generating from the left-most token to the right (left-to-right), while $\xleftarrow{}$ indicates going right-to-left. For Language Models (LMs), this means \lrLM is what is often called a ``forward LM'', while \rlLM is a ``backward LM''. For our sampling function (RD), 
this also indicates which generated context is being conditioned on, e.g. \lrRD conditions on left context, extending it to the right to generate output.

\subsection{Overview} 
\label{subseq_overview}

Currently, LM-decoding is limited to a \textit{text continuation} paradigm. Given an input text $s_{input}$, $\text{LM}(c|s_{input})$ generates contexts $c$ that might come after (forward, i.e. \lrLM) or before (backward, i.e. \rlLM) the input. LM-decoding generates \textit{outside} of the input by continuing it, but many tasks require us to generate \textit{over} or \textit{within} the input: paraphrasing requires reformulating the input, while infilling requires inserting text in the middle of it. 

Reflective Decoding approaches this shortcoming by turning conventional LM-decoding around. While $\text{LM}(c|s_{input})$ generates the kinds of contexts $c$ the input might appear in, $\text{RD}$ generates $s$ that might replace $s_{input}$ in these same contexts. The distributional hypothesis \citep{firth1957synopsis} suggests semantically similar texts appear in similar contexts, meaning $\text{RD}$ is also likely to sample in the semantic neighborhood of $s_{input}$.

Concretely, \method samples $s$ that fits the same contexts as $s_{input}$ in 2 simple steps.
We first sample many representative
contexts $c_i$ that could neighbor the input, e.g. using \lrLM in Figure \ref{fig_1}. This is the \textbf{\stepone}. Second, in the \textbf{\steptwo}, we generate text in the opposite direction (using \rlLM in Figure~\ref{fig_1}), which fits these contexts as well as $s_{input}$ fits them. To consider all $c_i$'s while decoding, we ensemble the different distributions imposed by conditioning on each $c_i$:

\input{figs/algorithm_simple}

\begin{equation}
    \text{\rlRD(s)} = \frac{ \prod_i \text{\rlLM}(s|c_i)^{w_i}}{Z(s,c,w) }
    \label{eq:RD_simple}
\end{equation}
\noindent
where $Z$ normalizes the fraction to a proper probability distribution (see Equation \ref{eqn:rd}). In essence, this ensemble \rlRD restricts generations to fit all contexts $c_i$. Reversed \lrRD is the same, except it uses \lrLM with \emph{left} contexts $c_i$ generated by \rlLM.

By ensembling the contexts in a Product of Experts \citep{hinton2002training} framework, we can generate a hypothesis $s$ that fits the full contextual fingerprint. Yet, some contexts are more informative than others:  probable but generic contexts like ``See the appendix for details.'' are not descriptive of neighboring text. We learn weights $w_i$ to prioritize contexts $c_i$ in the ensemble that are most informative for $s_{input}$, by maximizing the probability of $s_{input}$ under Equation~\ref{eq:RD_simple} (described in Algorithm~\ref{alg}). In effect, we are learning an on-the-fly autoencoder at inference time, using weighted ensembles of contexts as a 
representation (see \S\ref{subseq_intuitions_and_theory}, \S\ref{app:derivation}).

To motivate how this method functions, consider the paraphrasing example from Figure~\ref{fig_1} with input $s_{input}=$ \textit{How are circulatory system tissues formed?} Generated contexts reflect different aspects of $s_{input}$: $c_1$ situates $s_{input}$ as a question (\textit{This is a medical question...}), while $c_2$ and $c_3$ explore central concepts  (\textit{as with all tissue...}; \textit{about the circulatory system}). Even though each context could follow many sentences, together they form a fingerprint for $s_{input}$. A sentence that could be followed by all of $c_1,c_2,c_3$ will likely be a question ($c_1$), about tissue formation ($c_2$), and the circulatory system ($c_3$), and generally occupy the same semantic neighborhood as $s_{input}$, e.g. \textit{How do circulatory systems form?}


In the case of paraphrasing, our task is to replace all of $s_{input}$ with something that might appear in the same contexts. Other tasks, however, might require us to replace only part of a sentence (e.g. in-context paraphrasing) or even insert text at a given position (e.g. infilling). \method makes this easy: simply hold part of $s_{input}$ static when we generate from RD.


\vspace{15pt}

\subsection{\method}
\label{subsec:reflective_decoder}

Here we dive into the details of \method, by considering the right-hand context ensemble (\rlRD), keeping in mind that the process is repeated on the left-hand as well (\lrRD).

First, in the \textbf{\stepone} (line 1 of Algorithm~\ref{alg}), we sample many right-hand contexts $c_i$ for $s_{input}$, using \lrLM. These will be used as a representative sample of the contexts $s_{input}$ appears in.
Second, in the \textbf{\steptwo} (lines 2 \& 3) our goal is to construct a sampling function \rlRD that will yield texts similar to $s_{input}$. We define \rlRD as:
\begin{equation}
    \text{\rlRD}(s) = \frac{ \prod_i \text{\rlLM}(s|c_i)^{w_i}}{ \prod_{j=0}^{|s|} \sum_{t \in V}  \prod_{i}  \text{\rlLM}(t|s_{j+1:|s|} + c_i)^{w_i} }
    \label{eqn:rd}
\end{equation}

\noindent
This is equivalent to Equation~\ref{eq:RD_simple}, but giving the exact normalization factor in the denominator. 

Equation~\ref{eqn:rd} is a token-wise Product of Experts model, that captures the semantic neighborhood of $s_{input}$ via the combination of contexts $c_i$ and their weights $w_i$ (\S\ref{subseq_intuitions_and_theory}). 
We learn $w_i$ that maximize $\text{\rlRD}(s_{input})$ (probability of generating $s_{input}$ under \rlRD), thereby up-weighting contexts specific to $s_{input}$. We initialize these weights (line 2),  then train them (line 3) using the Adam optimizer \citep{kingma2014adam}. We normalize weights into a proper probability distribution at every step. 

Reverse-direction \lrRD is learned symmetrically, flipping the roles of \lrLM and \rlLM and sampling left-hand context instead (see \S\ref{app:lr} for details). Finally, we generate from \rlRD (and \lrRD), sampling outputs that would appear in the same contexts as $s_{input}$. Depending on the application, we rank and select a final output in different ways, always using \lrLM and \rlLM together to capture bidirectional fit.

\input{figs/fig_examples}

\subsection{Implementation}
\label{subsec:implementation}


\paragraph{Weight Learning and Pruning} Context weights $w_i$ are learned using the Adam optimizer \citep{kingma2014adam}. In practice this takes under 100 steps (negligible time compared to LM decoding). While we sample tens of contexts (line 1 of Algorithm~\ref{alg}), many end up with negligible weight under the learned distribution (Equation~\ref{eqn:rd}). To efficiently sample from \rlRD and \lrRD, we drop all but the top $k_c$ contexts and renormalize weights: $k_c < n_c$ contexts are used during the \steptwo.

\paragraph{Parameters} We sample $n_c$ contexts to describe the source $s_{input}$. We use nucleus sampling \citep{Holtzman2019TheCC} with parameter $p_c$, and a maximum length of $len_c$. 
Once \lrRD and \rlRD are learned, we sample $n_{s}$ generations from each, of length $len_{s}$. We again use nucleus sampling, but choose $p_{s}$ per-example to account for vastly different entropy in RD (\S\ref{app:entropy_calibtration}). Values for all hyperparameters are available in \S\ref{app:model_params}.

\paragraph{Language Models} We train large forward (\lrLM) and backward (\rlLM) Language Models based on GPT-2 \citep{radford2019gpt2}  using the OpenWebText training corpus \citep{Gokaslan2019OpenWeb}\footnote{https://github.com/yet-another-account/openwebtext}. Our implementation details follow those of past work retraining GPT-2 \citep{zellers2019defending}.

\subsection{Application: Paraphrasing}
\label{subsec:app_paraphrase}

To paraphrase, we begin by generating candidate outputs. Following \S\ref{subsec:reflective_decoder} the \decoder is learned in each direction (\lrRD, \rlRD) using the source sentence $s_{input}$. Then, $n_{s}$ generations are sampled from both \lrRD and \rlRD: \
$$s_1, ..., s_{n_s} \sim \text{\lrRD}, s_{n_s + 1}, ..., s_{2*n_{s} } \sim \text{\rlRD}$$

\noindent
This gives a robust set of candidates that are compatible with the same left and right contexts as $s_{input}$. Many of these will be semantically related to $s_{input}$, but must be scored and ranked in order to select true paraphrases.
\method is based on the notion that good ``fit'' with the same contexts is a robust measurement of similarity, yielding a natural ``contextual scoring function'' (Equation \ref{eqn:reflective_scoring} and \S\ref{subseq_intuitions_and_theory}). We measure how likely candidate $s$ is to generate the same contexts that $s_{input}$ did when constructing \lrRD and \rlRD:
\begin{equation}
    score(s) = \frac{1}{n_c}\sum_{c_{rh}} \text{\lrLM}(c_{rh}|s) + \frac{1}{n_c}\sum_{c_{lh}} \text{\rlLM}(c_{lh}|s)
    \label{eqn:paraphrasing_full_score}
\end{equation}
\\
\noindent
where $c_{rh}$ are the generated contexts used in \rlRD, and $c_{lh}$ for \lrRD.
This explicitly estimates how similar the contexts of $s$ and $s_{input}$ are on both sides, the underlying objective of \method.


\subsection{Application: Abductive Reasoning}
\label{subsec:app_anlg}

Abductive natural language generation ($\alpha$NLG from \citealt{Bhagavatula2019AbductiveCR}) is the task of filling in the blank between 2 observations $o1$ and $o2$, with a hypothesis $h$ that abductively explains them. The challenge for LM-decoding is making use of context from both sides ($o_1$ on the left and $o_2$ on the right). This is particularly challenging for unsupervised decoding methods because unidirectional LMs cannot naturally condition on both sides when generating $h$. 

\method simplifies this problem by capturing information about both $o_1$ and $o_2$ in a single decoding function (\rlRD or \lrRD), then holding $o_1$ and $o_2$ static at generation time (i.e. teacher forcing).
Concretely, we use concatenated $o_1 + o_2$ as $s_{input}$ in Algorithm~\ref{alg}, and construct sampling functions \lrRD, \rlRD informed by both observations. 
We are interested in sampling in between $o_1$ and $o_2$, so when sampling hypotheses $h$ from \rlRD we condition on the right-side observation $o_2$ (and vice-versa for \lrRD and $o_1$). This is equivalent to appending the given observation to sampled contexts: 
\begin{equation}
\begin{split}
    h_1, ..., h_{n_{\hat{s}}} &\sim \text{\rlRD}(h|o_2) \\ \hspace{4 pt} h_{n_{\hat{s}} + 1}, ..., h_{2*n_{\hat{s}}} &\sim \text{\lrRD}(h|o_1)
\end{split}
\end{equation}

Note that both \lrRD and \rlRD contain information about \textbf{both} $o_1$ and $o_2$, effectively turning a 2-sided contextual constraint into a 1-sided one. \ari{``turning'' here is a bit strange, what if we said something about \textit{encoding} information from $o_1$ and $o_2$ or some such}

We also use a task-specific scoring function to rank sampled hypotheses. \ari{the previous sentence sounds weird, I would just mention ranking in the following sentence and delete it} We would like a hypothesis $h$ that best explains both observations, and so use Language Models to measure this:
\begin{equation}
    score(h) = \text{\rlLM}(o_1| h 
+ o_2) + \text{\lrLM}(o_2| o_1 + h )
\label{eqn:anlg_score}
\end{equation}

Adding $h$ should help to ``explain'' each observation given the other, i.e. that $o_2$ follows from $o_1 + h$ and $o_1$ from $h + o_2$. To filter hypotheses that only explain one of the two observations, we remove any that make either observation less probable than the empty hypothesis, imposing:
\noindent
\begin{align*}
    \text{\rlLM}(o_1|h + o_2)>\text{\rlLM}(o_1|o_2) \\
    \text{\lrLM}(o_2|o_1 + h)>\text{\lrLM}(o_2|o_1)
\end{align*}
\subsection{Intuitions and Theory}
\label{subseq_intuitions_and_theory}
Here we discuss the theoretical intuition for \method, as a way to sample generations that share contextual ``fit'' with a source text, deriving the sampling function of Equation~\ref{eqn:rd}. 

We start by considering how to relate the meaning of two texts, generation $s$ and input $s_{input}$. We follow a distributional intuition \citep{firth1957synopsis}, that meaning can be understood through the contexts in which text appears. Many distributional approaches learn contentful neural representations by predicting context given input text \citep{word2vec,kiros2015skip}, then compare these representations to establish semantic similarity. We can, instead, compare contexts directly—judging the difference in meaning between texts $s_{input}$ and $s$ by their divergence:
\begin{equation}
    D_{KL}(\text{\lrLM}(c|s_{input}), \text{\lrLM}(c|\hat{s}))
    \label{eqn:context_divergence}
\end{equation}
We use \lrLM to interchangeably denote the theoretical left-to-right distribution of text, and the LM estimating it. Thus, $\text{\lrLM}(c|s)$ is the distribution over right contexts $c$ given sentence s, and Equation~\ref{eqn:context_divergence} can be understood as the ``contextual information difference'' we expect $s$ to have from $s_{input}$. Note, we could similarly use left-hand context and \rlLM—and do so in practice. 

We use finite-sample cross entropy as an effective empirical proxy for $D_{KL}$:
\begin{equation}
\begin{split}
 \hat{H}(\text{\lrLM}(c|s_{input}), & \text{\lrLM}(c|s)) = \\ &\frac{1}{N}\sum_{c_i \sim \text{\lrLM}(c|s_{input})} -\text{log}\text{\lrLM}(c_i|s) 
 \end{split}
 \label{eqn:reflective_scoring}
 \end{equation}
Where $c_i \sim \text{\lrLM}(c|s_{input})$ indicates sampling contexts for $s_{input}$ from \lrLM.
Intuitively, we want to minimize this score when generating $s$: an optimal output has a similar meaning to $s_{input}$ and so fills approximately the same contextual hole, minimizing the value of this ``contextual distance''. 

In this form, $\hat{H}$ compares 2 complete texts--$s$ and $s_{input}$--but we are trying to \textbf{generate} $s$ for which the divergence from $s_{input}$ is low. We flip the role of ``text'' and ``context''
\footnote{Context is a symmetric relation: a given text serves as the one-sided context of its own context.} 
to define a function from which we can sample $s$:
\begin{equation}
    \begin{split}
    &\text{\rlRD}(s_j|, s_{j+1:n}) = \frac{ \prod_i  \text{\rlLM}(s_j|s_{j+1:n} + c_i)^{w_i}}{ \sum_{t \in V}  \prod_i  \text{\rlLM}(t|s_{j+1:n} + c_i)^{w_i} }
    \end{split}
    \label{eqn:rd_theory}
\end{equation}
(equivalent to Equation~\ref{eqn:rd}, derived in \S\ref{app:derivation}) $s_j$ is the $j^{th}$ token in $s$ (sampled right-to-left from n to 0), and $V$ is the vocabulary. Weights $w_i$ are learned
by maximizing the probability of $s_{input}$. 

Equation~\ref{eqn:rd_theory}, estimates the probability of \textit{predicting} $s_{input}$ and $s$ from a finite set of contexts $c_i$ generated from $s_{input}$. This approximately minimizes Equation~\ref{eqn:context_divergence}, as being generated by the same weighted ensemble of contexts strongly correlates with generating the same contexts in the same proportions, i.e. low divergence, due to the sparsity of language.
We can sample $s$ with low contextual distance from $s_{input}$ using \rlRD. Further, we can use left context to construct \lrRD by simply reversing the directions of the LMs used.


%% file: figs/algorithm_simple.tex

\begin{algorithm}[!t]
\small
\centering
\caption{\scriptsize Learn \rdexp $\overleftarrow{\text{RD}}$}
\label{alg}
\begin{algorithmic}[1]
\REQUIRE Forward language model \lrLM \\
\hspace{1.2em} Backward language model \rlLM
\\ \hspace{1.2em} Source text $s_{input}$

\SetAlgoLined

\STATE  Sample contexts, $c_1...c_{n_{c}} \sim \text{\lrLM}(c|s_{input})$\\
\STATE Initialize parameters $\mathbf{w} = w_1...w_{n_{c}}$ s.t. $\sum w_i = 1, w_i\geq0$\\
\STATE learn $\mathbf{w}$ = $\arg\max_{\mathbf{w}} \overleftarrow{\text{RD}}(s_{input})$ \\s.t. $\sum w_i = 1, w_i\geq0$\\
\ENSURE $\overleftarrow{\text{RD}}$
\end{algorithmic}
\end{algorithm}

%% file: figs/fig_examples.tex
\begin{figure*}[t]
    \centering
    \includegraphics[width=\linewidth]{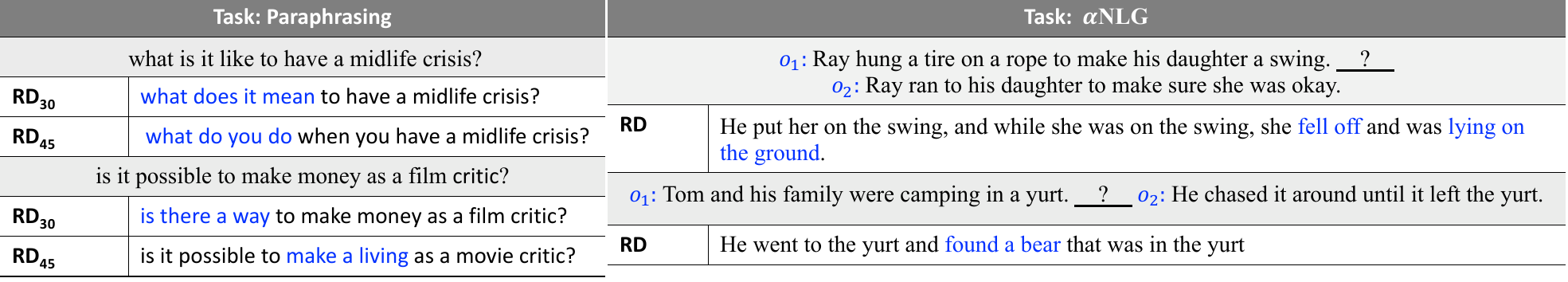}
    \caption{
    Example generations of \method on paraphrasing and abductive text infilling ($\alpha$NLG). $RD_{45}$ encourages more difference from the input than $RD_{30}$ (\S\ref{subseq_quora_gen}).
    }
    \label{fig_examples}
\end{figure*}

%% file: sections/experiments.tex
\section{Experiments}
\label{sec_experiments}

\input{figs/quora_results}

\subsection{Task: Paraphrase Generation} \label{subseq_quora_gen}

\paragraph{Task:} Following past work, we test our paraphrasing method (\S\ref{subsec:app_paraphrase}) on the \href{https://www.quora.com/q/quoradata/First-Quora-Dataset-Release-Question-Pairs}{Quora question pair dataset}. 
We hold out 1000 examples for testing, with the rest for training and validation (used by supervised baselines), disallowing overlap with the test set. We test a subset of models (compatible unsupervised models, MT) on the Twitter URL corpus \citep{lan2017continuously}, using 1000 examples from the canonical test split. 

\paragraph{Metrics:}
Following past work, we include automatic metrics BLEU \citep{papineni2002bleu}, METEOR \citep{denkowski2014meteor}, and $\text{TER}_p$ \citep{snover2009ter}. These measure agreement with references, but high reference/input overlap means copying is rewarded  \citep{Mao2019PollyWA}; indeed, \emph{copying source sentences as-is wins on these metrics (Table~\ref{tab:quoratest}), meaning both BLEU and METEOR can be easily gamed.} 

Past work has emphasized the important challenge of generating novel paraphrases \citep{liu2010pem,chen-dolan-2011-collecting}
We address this in 3 ways. First, we explicitly quantify a simple notion of novelty: 
\begin{equation}
    Novelty(\hat{s}) = 100 - BLEU(\hat{s}, s_{input})
    \label{eqn:novelty}
\end{equation}
\noindent 
to quantify the novelty-quality trade-off. Second, we include the SARI metric \citep{xu-etal-2016-optimizing} which explicitly balances novelty from input with reference overlap. Third, we quantify an overall human quality metric accounting for this. 

We have humans evaluate fluency, consistency, and novelty on Amazon Mechanical Turk. The overall score (``Human'' in Table~\ref{tab:quoratest}) \ari{just saying ``Human'' when the other things here are from humans is confusing} is the rate examples meet thresholds for all 3: fluent enough to understand, with at most minor differences in meaning and at \emph{least} minor differences in wording. On quora, we test 200 examples, with agreement  (Fleiss' $\kappa$ \citealp{fleiss1971measuring}) of 0.40 (fluency), 0.54 (consistency), 0.77 (novelty) and 0.48 (overall) i.e. moderate to substantial agreement \citep{landis1977measurement}. On the Twitter corpus, we use 100 examples with agreement of 0.39, 0.42, 0.54, and 0.36, indicating fair to moderate agreement. On both we have 3 raters per example. See \S\ref{app:human_metrics} for more.

\paragraph{Baselines:}
Parameters for \method are given in \S\ref{app:model_params}. We mainly compare against 3 unsupervised baselines: Controlled Sentence Generation by Metropolis Hastings (CGMH from \citealt{miao2019cgmh}), Simulated Annealing (UPSA from \citealt{liu2019upsa}) and the residual VQ-VAE of \citet{roy2019unsupervised} (R-VQVAE). This is a cross-section of recent approaches (VAE, editing).

We also compare against a machine-translation approach (see Sec \ref{sec_related}), pivoting through German using Transformer \citep{vaswani2017attention} models trained on WMT19 data \citep{WMT2019}.
MT is included in a separate section in our results as it uses supervised bilingual data (Table~\ref{tab:quoratest}).

We include supervised baselines: the pointer generator trained by imitation learning (PG-IL) as in \citet{WYU2019}, the diversity-promoting DiPS model \citep{DiPS2019}, and a finetuned BART model \citep{lewis2019bart}, which uses a more complex pretraining method than our LMs.  Note that DiPS generates multiple diverse paraphrases so we pick one at random. 

CGMH and \method both return multiple sampled, ranked paraphrases. We can easily control for $Novelty$ by taking the highest-ranked output that meets a $Novelty$ threshold. For both, we have a version with no threshold ($Top$), and with thresholds such that average $Novelty$ is $30$ and $45$. $Novelty$ cutoffs do not depend on the reference, only the source, and are equivalent to selecting with \textit{BLEU-ori} ($Novelty$ is $100-\text{BLEU-ori}$) by \citet{miao2019cgmh} or \citet{bao2019dss}.

\subsection{Task: Abductive NLG}
\label{subseq_alpha_nlg}

\paragraph{Task:}
The Abductive natural language generation task ($\alpha$NLG) presented in \citet{Bhagavatula2019AbductiveCR} requires generating a hypothesis that fits between observations $o_1$ and $o_2$, and explains them. We apply \method to this problem as outlined in \S\ref{subsec:app_anlg}, using the given data splits.

\paragraph{Metrics:}
For human evaluation, over 200 examples we ask 3 raters on Amazon Mechanical Turk about coherence between h and $o_1$, $o_2$, $o_1+o_2$, and overall quality on 4-value likert scales. We found Fleiss' kappa \citep{fleiss1971measuring} of 0.32, 0.40, 0.41, and 0.41 respectively, indicating fair to moderate agreement \citep{landis1977measurement}.

\paragraph{Baselines:} Parameters for \method are given in \S\ref{app:model_params}.  We include baselines from the original work: different supervised variants of GPT-2 large with access to the observations, and optionally commonsense embeddings or generations from COMET \citep{bosselut2019comet}. We include unsupervised baselines of GPT-2 conditioned on $o_1$ + $o_2$ directly, the gradient-based DeLorean model of \citet{qin-etal-2020-back}, and ILM infilling model of \citet{donahue-etal-2020-enabling}, representing recent unsupervised methods. 

%% file: figs/quora_results.tex
\begin{table*}[t]
\small
    \centering
    \begin{tabular}{ll|ccccc|c}
     \toprule
 & Method & SARI$\uparrow$ & BLEU$\uparrow$ & METEOR$\uparrow$ & TER$_{P}$$\downarrow$ & Human$\uparrow$ &  $Novelty\uparrow$ \\ 
 \midrule
 \textit{Human} & Source & 17.8 & 56.0 & 37.6 & 48.0 & - & 0.0 \\
   & Reference & 91.9 & 100.0 &100.0 & 0.0 & 71.7 & 43.9 \\ 
 \midrule
\textit{Supervised} & PG-IL & 32.8 & \textbf{49.1} & 33.8 & \textbf{49.0}* & 29.4 & 24.4 \\
   & DiPS & \textbf{38.8} & 41.0 & 27.9 & 56.0 & 36.6 & \textbf{48.5}* \\ 
   & BART & 36.1 & 44.7 & \textbf{34.7}* & 66.0 & \textbf{46.1} & 35.2 \\ 
 \midrule
   \textit{Supervised (Bilingual)} & MT & 35.6 & 48.1 & 33.5 & 52.0 & 59.3 & 26.8\\
   \midrule
  \textit{Unsupervised} & R-VQVAE & 27.2 & 43.6 & \underline{32.3} & 60.0  & 33.5 & 26.2\\ 
   & CGMH$_{Top}$& 32.3 & 42.0 & 28.2 & 59.0 & 27.0  & 27.6\\ 
   & CGMH$_{30}$ & 33.9 & 40.9 & 27.5 & 60.0  & 31.5 & 29.7\\ 
   & CGMH$_{45}$ & 32.6 & 33.8 & 23.4 & 65.0  &  15.8 & \underline{44.5}\\ 
   & UPSA & 34.0 & 36.6 & 26.7 & 70.0 & 37.8 & 44.4\\
   \midrule
  & RD$_{Top}$ (Us) & 29.0 & \textbf{49.9}* & \textbf{33.9} & \textbf{52.0}  & 27.5 & 20.8\\
   & RD$_{30}$ (Us) & \textbf{40.0}* & \underline{46.8} & 32.2 & \underline{57.0} & \textbf{63.2}* & 30.0\\
   & RD$_{45}$ (Us) & \underline{38.6} & 39.9 & 28.9 & 65.0 & \underline{61.1} & \textbf{45.0}\\
    \bottomrule
    \end{tabular}
    \caption{Model performance on the Quora test split. \textbf{Bold} indicates best for model-type, * indicates best overall (excluding human), \underline{underline} indicates second-best for unsupervised. The first 5 columns are measures of quality, while the last measures novelty (Equation~\ref{eqn:novelty}) or difference from the input. We rerun evaluations from past work. }
    
    \label{tab:quoratest}
\end{table*}

%% file: sections/results.tex
\section{Results and Analysis}
\label{sec_results}
\label{subsec_discussion}

\input{figs/twitter_results_small}

\paragraph{Paraphrasing} 
First, the Quora dataset: On automatic metrics from past works (BLEU, METEOR, \TERP) our lowest-$Novelty$ model setting ($\text{RD}_{Top}$) achieves the highest unsupervised scores, and highest overall on BLEU. Other high scoring rows (Source, PG-IL) are similarly low-$Novelty$. The SARI metric explicitly balances $Novelty$ with similarity to reference. On SARI we see such low-$Novelty$ models perform worse. The best overall model on SARI is our medium-$Novelty$ setting ($\text{RD}_{30}$) which outperforms MT and supervised models. 

Our human evaluation measures what fraction of outputs are found to be fluent, consistent, and novel. As with SARI, both our mid and high-$Novelty$ models perform quite well, again with the medium-$Novelty$ setting outperforming all baselines. As further validation for SARI as a proxy for human, they share the same top-5 models. 

Results on the Twitter URL corpus largely support those on Quora. \method achieves the best unsupervised scores on novelty-aware metrics (Table~\ref{tab:twitter_results_small}), with the best overall SARI, even outperforming reference on the human metric, although MT achieves the highest overall.

In sum, \method is able to compete on previously used quality metrics favoring low-$Novelty$, but can produce more varied outputs preferred by humans. $\text{RD}_{45}$ is among the best models by SARI and Human on Quora despite exceeding the novelty of even the reference.

\paragraph{$\alpha$NLG}
Results on $\alpha$NLG (Table~\ref{tab:anlg})
present a strong case that \method{} can effectively use bidirectional context. Strong hypotheses use information from both initial the observation $o_1$ and the future observation $o_2$. Humans ranked the ability of \method{} to capture this 42.4, about 17 points above the next-best unsupervised baseline and only 15 points below the best supervised method tested. We see similar results for overall evaluation. 
A likely factor in this is the (comparatively) high degree of coherence between $h$ and $o_2$ by \method. Where other methods seem to pay more attention to observation $o_1$ (the $o_2$ column generally has much lower values), \method has comparably high coherence with left-hand ($o_1$) and right-hand ($o_2$) contexts.

We also include example generations in Figure~\ref{fig_examples} to demonstrate the ability of \method to combine $o_1$ and $o_2$. For example, $h=$ \textit{He put her on the swing, and while she was on the swing, she fell off and was lying on the ground.} incorporates information from both observations. Specifically, it takes into account the swing that Ray is building for his daughter which is only mentioned in $o_1$, and hypothesizes about a potential injury due to Ray checking on his daughter in $o_2$. See appendix for more generations.

Overall, the strong performance of \method on $\alpha$NLG shows that unsupervised generation with context ensemble applies to infilling in addition to paraphrasing.

%% file: figs/twitter_results_small.tex
\begin{table}[t]
\small
    \centering
    \begin{tabular}{l|cc|c}
     \toprule
 Method & SARI$\uparrow$ & Human $\uparrow$ &  $Novelty\uparrow$ \\ 
 \midrule
 Source & 13.6  & - & 0.0\\ 
 Reference & 90.7 & 51.3 & 63.3 \\ 
 \midrule
 MT & 36.1 & 70.9 & 30.4\\
   \midrule
  R-VQVAE & 31.1 & 32.3 & 40.4\\ 
   CGMH$_{Top}$&  32.7 & 27.8 & 25.5\\
   CGMH$_{30}$&  33.2 & 25.1 & 30.1\\
   CGMH$_{45}$&  31.8 & 13.5 & 45.2\\
  RD$_{Top/30}$ (Us) & 31.4 & 46.5 & 37.0\\
   RD$_{45}$ (Us) & \textbf{36.4} & \textbf{56.9} & \textbf{45.3}\\
    \bottomrule
    \end{tabular}
    \caption{Model performance on the Twitter URL test split. \textbf{Bold} indicates best for model-type. We show only metrics accounting for novelty (more in \S\ref{subsec_twitter})}
    
    \label{tab:twitter_results_small}
\end{table}

%% file: sections/discussion.tex
\section{Discussion}

\paragraph{\method Out-of-the-Box}
A major advantage to applying \method is ease-of-use: armed with our pretrained language models, practitioners can immediately begin generating. With general pretrained models and underlying principles that are domain-agnostic, \method works across a broad range of text style--no finetuning required--making exploration and adaptation simple. Multiple rounds of generation mean \method{} may run slower than other methods at inference time\footnote{Depending on parameters we found most baselines took multiple seconds per example vs. 10s of seconds for \method on a multi-gpu machine.}, but it avoids training time. There are clearly settings that favor supervised learning (narrow, known domain with abundant training data), but \method is a good option to begin generating and exploring immediately with high quality generation. 

A useful abstraction for understanding \method for current applications is ``prompting'', i.e., writing a prefix to implicitly or explicitly describe a task for a pretrained model. \method generates natural contexts that the desired generation would appear in. This breaks from other methods of automatic prompting, which often forego ``natural'' prompts \cite{shin-etal-2020-autoprompt, reynolds2021prompt}, even making them continuous \cite{Li2021PrefixTuningOC, hambardzumyan2021warp, lester2021power, qin2021learning}. \method{} also notably creates a set of prompts (contexts) for each example, where other methods attempt to learn an overall task prompt. Still, all of these are connected by the popular intuition that useful behavior in pretrained models can be induced through contextual input.

\paragraph{Future Applications}

\method can extend beyond our experiments here, however. A simple example is in-context paraphrasing, i.e. writing a paraphrase that fits the true context that the original sentence appears in. Most existing paraphrasing methods consider only out-of-context sentences, and would require significant changes to consider context as a constraint; for \method we can simply combine true and generated contexts without with the same algorithm. 

Driving \method is a notion of context as a representation, with clear potential for future work. Pretrained LMs capture rich information about text spans, but accessing it without fine-tuning is nontrivial; within the model it is an uninterpretable mass of parameters and activation weights. Our work observes that unidirectional LMs are only capturing this information to predict adjacent context--this is the sole learning signal--so all of this information is expressed in the model's context prediction. Thus, we capture some of this rich information to represent spans, by capturing a finite-sample version of this full predictive distribution in generated contexts. In \method specifically, we use this form of representation to generate back into the source span--paraphrasing or infilling--but the notion can be applied much more generally. In translation for instance, we might first generate contexts for the source sentence that represent its meaning, noisily translate these contexts, then impose that any translations for the source fit the same contexts under a translation-language LM. Constraining translations in this way can add robustness to existing systems by anchoring translations to informative contexts. Beyond explicit generation even, we might use a very large LM like GPT-3 to define a strong scoring function or metric as in Equation~\ref{eqn:reflective_scoring}, first generating contexts for some target sentence, then scoring candidates by how well they generate these same contexts. As in our work, such a score could indicate how well the option fills the same contextual role as the target, harnessing the strong reasoning of whatever model is used.

%% file: sections/background.tex
\section{Related Work}
\label{sec_related}


\paragraph{Distributional Intuitions}
A key aspect of \method is using a distributional intuition to represent the meaning of a text through many contexts. \citet{kiros2015skip, miao2019cgmh} quantify semantic relationships and \citet{lin2001dirt} identify paraphrastic relationships under similar intuitions. A major point of difference between past work and ours is that we sample explicit contexts, allowing unsupervised generation back from these contexts, while past work typically learns a neural representation based on contexts and conditions on this vector-encoded representation.

\input{figs/anlg_results}

\paragraph{Unsupervised Paraphrasing}

Some approaches train neural variational auto-encoders unsupervised to represent source sentences, then decodes from these representations to paraphrase \citep{Roy2019,bao2019dss}. This requires training specialized representations, whereas \method applies general-purpose LMs. We compare to \citet{Roy2019}. 

Paraphrasing by editing the input \citep{miao2019cgmh, liu2019upsa} has shown promise. Like \method, these approaches can be applied without training specialized models,
but are necessarily limited by edit-paths and local minima, as edits are often restricted to single-word replacement, insertion, and deletion. Generated paraphrases must follow a continuous local edit path, while \method can generate new sentences from scratch.

\method and MT-based paraphrasing both pivot through an alternative textual form to paraphrase (context and translation, respectively). But MT paraphrasing systems cycle-translate through a pivot language \citep{federmann-etal-2019-multilingual, wieting2018paranmt}, which requires supervised bilingual translation data, with an implicit notion of interlingual paraphrasing.

\paragraph{Novelty in Paraphrasing}
\citet{Mao2019PollyWA} observe that paraphrases close to the source often win on automatic quality metrics. However, dissimilarity from the source correlates with human notions of paraphrasing \citep{liu2010pem}. \citet{DiPS2019} increase novelty through their diversity-promoting sampling method. Alternative metrics that consider novelty alongside quality have been proposed \citep{sun-zhou-2012-ibleu, federmann-etal-2019-multilingual}. The SARI metric \citep{xu-etal-2016-optimizing}, included here, combines these notions. 

\paragraph{Abductive Text Infilling} $\alpha$NLG \cite{Bhagavatula2019AbductiveCR} is a text infilling task that specifically measures the ability of models to \textit{explain} bidirectional context (observations $o1,o2$) with a hypothesis that fits between them. This naturally fits \method, which fills in contextual gaps. Recent work has directly addressed this task \citep{qin-etal-2020-back} while the infilling literature is also quite applicable \citep{donahue-etal-2020-enabling}. We compare to both of these methods on abductive infilling, showing superior results.

%% file: figs/anlg_results.tex
\begin{table}[t]
\small
    \centering
    \begin{tabular}{l|ccc|c}
     \toprule
       & $o_1$ & $o_2$ & $o_1+o_2$ & all\\ 
       
    \midrule  
    Human & 86.3 & 89.1 & 85.1 & 84.4 \\
   \midrule
   \rowcolor[gray]{0.95} \multicolumn{5}{l}{\textit{Supervised}} \\
   \midrule
   $\text{COMeT}_{\text{Emb}}$+GPT2 & \textbf{69.3} & \textbf{60.1} & \textbf{56.4} & \textbf{56.3} \\ 
   $\text{COMeT}_{\text{Txt}}$+GPT2 & 68.9 & 54.8 & 51.9 & 50.6 \\ 
   $O_1$-$O_2$-Only & 69.2 & 57.7 & 54.3 & 53.8  \\ 
    \midrule
   \rowcolor[gray]{0.95} \multicolumn{5}{l}{\textit{Unsupervised}} \\
   \midrule
   GPT2-Fixed & 20.6 & 13.9 & 10.8 & 10.3 \\ 
   DeLorean & 48.7 & 24.6 & 23.6 & 22.5 \\
   ILM & 45.9 & 27.3 & 25.3 & 25.0 \\
   Reflective Decoding & \textbf{53.4} & \textbf{51.7} & \textbf{42.4} & \textbf{41.9} \\

     \bottomrule
    \end{tabular}
    \caption{Model performance on $\alpha$NLG. The first 3 scores query agreement between hypothesis and given observation(s), ``all'' indicates overall judgement. \method significantly outperforms all unsupervised baselines.}  
    \label{tab:anlg}
\end{table}

%% file: sections/conclusion.tex
\section{Conclusions}
\label{conclusion}
We present \method, a novel unsupervised text generation method for tasks that do not fit the text continuation paradigm. It uses just two pretrained Language Models to generate contexts that capture aspects of input text, generating back into the input from there. It significantly outperforms unsupervised baselines in quality and novelty for paraphrasing. Further, in abductive natural language generation it outperforms unsupervised baselines by a significant margin and halves the gap with supervised models. \method uses the concept of representing meaning with generated contexts, offering new possibilities for unsupervised conditional text generation.

%% file: sections/acknowledgements.tex
\section*{Acknowledgements}
We thank anonymous reviewers for many helpful comments. 
This research is supported in part by the Natural Sciences and Engineering Research Council of Canada (NSERC) (funding reference number 401233309), DARPA CwC through ARO (W911NF15-1-0543), DARPA MCS program through NIWC Pacific (N66001-19-2-4031), the Allen Institute for AI, and a gift from Intel Labs Cognitive Computing Research.

%% file: sections/broader_impact.tex
\section*{Ethical Considerations}
\label{broader_impact}
In order to complete our human evaluation we used Amazon Mechanical Turk. We estimated the range of times we expected our task to take, and made sure that at minimum workers would be paid a wage of \$15.00 per hour if they were solely completing our task.

As part of this effort, we plan to release our code and model. Our forward and backward language models are the same size as the publicly available GPT-2~\cite{radford2019gpt2}. Training time/energy was likely significantly smaller than the original release; existing code and hyperparameters were available, and we use a smaller dataset. Further, there is no publicly available backward GPT-2 model that we are aware of, so releasing a pair of forward and backward models that were trained on the same data allows for proper comparisons about left-to-right vs. right-to-left processing of English text.

We estimate that the potential dangers of releasing this from a malicious generation perspective are low. Our forward model is similar to already released GPT-2 models. While the backward model adds new generation potential and scientific novelty, it is unlikely to compare to GPT-3~\cite{Brown2020LanguageMA} which many hobbyists and private companies now have access to. 
We believe that releasing a pair of forward and backward models will be more useful to researchers who wish to study the symmetries and asymmetries of the linguistic distribution.

%% file: sections/appendix.tex
\section{Appendix}

\subsection{Derivation of Sampling Function}
\label{app:derivation}


Here we derive the sampling function used for \method, which allows generation using contextual similarity. This supplements \S\ref{subseq_intuitions_and_theory}. $P_{c|s}$ denotes the distribution of contexts $c$ for sentence $s$. This will be 1-sided context, for instance right-hand context $c_{rh}$ (i.e. $P_{c|s}$ would be estimated by left-to-right \lrLM conditioned on s $\text{\lrLM}(c|s)$). Reversed $P_{s|c}$ goes back from context \textit{towards} text. With right-hand context, this is estimated by $\text{\rlLM}(s|c)$.




In \S\ref{subseq_intuitions_and_theory}, we consider the task of comparing a source sentence $s_{src}$ with another sentence $s$. For instance, we may want to know if $s$ is a paraphrase of $s_{src}$. Following a distributional intuition \citep{firth1957synopsis} we define a simple way to compare meaning:
\begin{equation}
    D_{KL}(P_{c|s_{src}},P_{c|s})
    \label{eqn:DKL_app}
\end{equation}
Where $D_{KL}$ is the Kullback–Leibler divergence measuring the difference between distributions $P_{c|s_{src}}$ and $P_{c|s}$. This captures a notion above: we take the amount the contexts of $s_{src}$ and $s$ differ as a proxy for their difference in meaning. 


In paraphrase generation, we want to select for contextual closeness, and thus only need to rank options. We will then use cross-entropy:
\begin{equation}
    \begin{split}
        H(\text{\lrLM}(c|s_{src}),&\text{\lrLM}(c|s)) \\
        &= \sum_{c} -\text{\lrLM}(c|s_{src})log(P_{c|s}(c))
    \end{split}
\end{equation}
\noindent
which is equivalent to $D_{KL}$ up to a constant offset, and is easier to estimate. Here, the sum over $c$ indicates every possible context $c$, but in practice we us finite samples. 

From Sec~\ref{subseq_intuitions_and_theory}, this quantifies contextual difference in meaning. For paraphrasing, we want a sentence $s$ that minimizes this, which is equivalent to maximizing the exponent of its negation:

\begin{equation}
    \begin{split}
        Score(s) &= e^{\sum_{c} -P_{c|s} log(P_{c|s}(c))} \\
        &= \prod_c  \left ( \frac{P_{s|c}(s) P(c)}{P(s)} \right ) ^{P_{c|s}(c)}  \\
        &= \frac{a_0}{P(s)} \prod_c  P_{s|c}(s)   ^{P_{c|s}(c)}  \\
    \end{split}
    \label{eqn:app_1}
\end{equation}
Constant $a_0$ results from factors of $P(c)$. 
The result is a Product of Experts \citep{hinton2002training}. $P(s)^{-1}$ will prioritize more context-specific paraphrases (low probability but likely in context). However, our LMs are not well equipped to handle unlikely text, (expressivity is likely spent on likely text). Second, while less likely text can have higher similarity, this may not be the goal of our system. Rather we want related sentences that are also \textit{fluent} and \textit{reasonable}, so we drop $P(s)^{-1}$, the equivalent of multiplying in $P(s)$, biasing the model towards likely sequences: 

\begin{equation}
    \begin{split}
        Score(s)
        &= c_0 \prod_c  P_{s|c}(s)^{P_{c|s}(c)}  \\
    \end{split}
\end{equation}

A product of experts of the form:

\begin{equation}
    \begin{split}
        Score(s) 
        &= \prod_c  P_{s|c}(s)^{w_{c|s}}  \\
    \end{split}
\end{equation}

We must set the weights $w_{c|s}$ in the finite sample setting. To keep in line with this the format, we will enforce that weights constitute a proper distribution. In the limiting case (unlimited samples) $w_{c|s}$ should be set to $P_{c|s}(c)$. However, these are likely not efficient estimation weights. Further, exponentiating by this estimate will magnify errors. Instead, we learn these weights using a heuristic, discussed later.

Next, we move to the finite-sample setting, replacing distributions with LM estimates. Here we will consider right-context (meaning $P_{s|c}$ is estimated by \rlLM) but the left-context case proceeds symmetrically. Substituting in the LM distribution:

\begin{equation}
    \begin{split}
        Score(s) 
        &= \prod_c  \text{\rlLM}(s|c)^{w_{c|s}}  \\
    \end{split}
    \label{eqn:app_PoE_score}
\end{equation}

Where now the product over $c$ indicates product over the finite sampled contexts. We convert this to a sampling function, decomposing into tokens of generation $s = s_0 ... s_n$:

\begin{equation}
    Score(s_{0:n}) = \prod_{j} \prod_c  \text{\rlLM}(s_j|s_{j+1:n})^{w_{c|s}}
\end{equation}

This restates equation~\ref{eqn:app_PoE_score} factorizing LM probability by tokens. Renormalizing and decomposing by token position gives a natural distribution to sample from:
\begin{equation}
    \begin{split}
            P_{sample}&(s_j|s_{j+1:n}) = \\ & \frac{ \prod_c  \text{\rlLM}(s_j|s_{j+1:n})^{w_{c|s}}}{ \sum_{t \in V}  \prod_c  \text{\rlLM}(t|s_{j+1:n})^{w_{c|s}} }
    \end{split}
    \label{eqn:rd_appendix}
\end{equation}
\noindent
normalizing token-wise over the vocabulary $V$ to a proper distribution (sampling right-to-left, index $n$ down, to match convention). This is referred to as \rlRD in the body of the paper, and stated in equation \ref{eqn:rd_theory}. This samples candidate generations that encourage adherence to the contextual scoring function. 

Finally, we learn the weights (a proper distribution): $s_{src}$ should receive the highest score (or similarly, should have the lowest contextual difference with itself, as it is likely in its own contexts).

\section{Implementation Details}
\label{app:implementation}

\subsection{Left-to-Right \decoder}
\label{app:lr}

From \S\ref{subsec:reflective_decoder}, \lrRD is learned similar to \rlRD, switching the roles of \lrLM and \rlLM in algorithm \ref{alg}. First, the roles of the language models are flipped in the sampling function:

\begin{equation}
    \text{\lrRD(s)}= \frac{ \prod_i \text{\lrLM}(s|c_i)^{w_i}}{ \prod_{j=0}^{|s|} \sum_{t \in V}  \prod_{i}  \text{\lrLM}(t|s_{0:j-1} + c_i)^{w_i} }
    \label{eqn:rd_lr}
\end{equation}
\noindent
$c_i$ are now generated by right-to-left \rlLM (i.e. left-contexts). see Algorithm~\ref{alg_lr}.

\input{figs/algorithm_lr}

\subsection{Post-processing Generations} Without learning stop-tokens, \method samples fixed number ($len_{s}$) of tokens. Candidates are extracted from raw generations using sentence tokenization.

\subsection{Entropy Calibration}
\label{app:entropy_calibtration}

Entropy calibration is used when sampling candidate generations (\S\ref{subsec:implementation}). When sampling output generations, generation parameters (truncation parameter $p_{s}$ from nucleus sampling, in paraphrasing) control how ``greedy'' or stochastic sampling is. However, the effect of $p_{s}$ depends on many dynamic (example-wise) factors. Setting $p_{s}$ too low may sample only the most likely option, too high gives off-topic candidates. The ``correct'' value of $p_{s}$ is highly example-dependent.

We define \textbf{entropy calibration} to control how much ``randomness'' is used in sampling in a robust way. Rather than directly setting a $p_{s}$ for all examples, this specifies the approximate entropy $\hat{h}$ to sample with for each example. In the greedy case for instance, the desired entropy $\hat{h}$ is set to 0 (i.e. picking from a set of 1 possible option).

We search for $p_{s}$ in each case that is expected to give the correct entropy for the full generation, although $p_{s}$ is a token-level parameter. To estimate this, we take sampling entropy over the source text $s_0...s_n$ under the nucleus-sampling truncated distribution P$_p$:
\begin{align}
 & \hat{h} = \\&\sum_i \sum_{w \in V_p} -P_p(w|s_0 ... s_{i-1}) \text{log} P_p(w|s_0 ... s_{i-1})  
\end{align}

\noindent
$V_p$ is the truncated vocabulary with parameter $p_{s}$. We select $p_{s}$ that gives a desired entropy, setthing this to 4 or 6 which we found effective (App.~\ref{app:model_params}).

\subsection{Parameters}
\label{app:model_params}
Here, we give model settings for our 2 experimental settings, paraphrasing and $\alpha$NLG. See Table~\ref{tab:params}. $\alpha$NLG requires higher variety (higher $h_{sample}$, $p_c$), and fewer generated contexts ($n_c$). We experimented with different reasonable values on the dev set of each model, evaluating manually. We use transformer language models (Mega size) trained on TPU pods (TensorFlow) of size 512. These will be made publicly available. For generation we used 2 NVIDIA Titan Xp GPUs.

\input{figs/model_params}

\section{Evaluation}

\input{figs/app_paraphrasing_examples}
\input{figs/app_anlg_examples}

\subsection{Automatic Metrics}
\label{app:auto_metrics}
Links to the automatic metrics: \href{https://pypi.org/project/pyrouge/}{ROUGE}, \href{https://www.nltk.org/_modules/nltk/translate/bleu_score.html}{BLEU}, \href{https://www.cs.cmu.edu/~alavie/METEOR/README.html}{METEOR}, \href{https://github.com/snover/terp}{\TERP}, \href{https://github.com/cocoxu/simplification/blob/master/SARI.py}{SARI}, \href{https://github.com/Tiiiger/bert_score}{BERTScore},\href{https://github.com/google-research/bleurt}{BLEURT}.
We include extra further metrics tested for Quora in table \ref{tab:quoratest_aux}: ROUGE \citep{lin-2004-rouge}, BLEURT \cite{Sellam2020BLEURTLR}, BERTScore \citep{Zhang2020BERTScoreET}. For BLEURT, and BERTScore we use default settings. We also include iBLEU \citet{sun-zhou-2012-ibleu} with $\alpha=0.9$.

\input{figs/quora_results_aux_metrics}

\input{figs/combined_human}

\subsection{Human Evaluation}
\label{app:human_metrics}
Human evaluation for Quora and Twitter are largely described in \S\ref{sec_experiments}. We reiterate that thresholds are used for each measure, and ``overall'' is the rate that all thresholds are met. Agreement is calculated on these binary combined threshold categories (following \citealt{combine_irr}). Full human results for paraphrasing are in Table~\ref{tab:combined_human}. Human eval for $\alpha$NLG is described in \S\ref{sec_experiments}.

\input{figs/quora_human}

\subsection{Twitter Dataset}
\label{subsec_twitter}
We include here the full results for paraphrasing on the Twitter URL corpus \cite{lan2017continuously}, a set of paraphrase pairs created by linking tweets with matching shared URLs. 
This marks a significant domain shift from our primary paraphrasing task (question paraphrasing). 
We test unsupervised models CGMH, R-VQVAE (UPSA Twitter model is not available), and the backtranslation MT model. R-VQVAE, the MT model, and \method do not use corpus-specific training data. CGMH trains on un-paired sentences from the Twitter training set, as outlined in the original work \cite{miao2019cgmh}. For all models, we use the same parameters as on the Quora dataset. 
We include supplementary results to the main paper in Table~\ref{tab:twitter_results}.

While we include SARI, Human, and $Novelty$ in the main paper, here we include results on a number of automatic metrics (table \ref{tab:twitter_results}). Results are similar to Quora: a novel setting of Reflective Decoding (in this case RD$_{45}$) achieves the highest score on SARI, which seemed most aligned with Human on Quora and is the only metric included here that accounts for both novelty and quality. The metrics that do not account for novelty are unsurpisingly dominated by generations with lower novelty: $RD_{Top}$ gets the highest unsupervised BLEU (MT is highest overall), while R-VQVAE gets the highest unsupervised METEOR and CGMH$_{Top}$ the best unsupervised TER$_P$. 

We also include full human results in table \ref{tab:twitter_human}.

\input{figs/twitter_results}

\input{figs/twitter_human}

\subsection{Reflective Scoring Evaluation}
\label{subsec_wmt18}
While the effectiveness of sampled contexts as a representation for paraphrasing is supported by our main experiments (\S\ref{sec_experiments}), we would like to provide further validation of the semantic capacity of generated contexts, and validate the underlying scoring function of \method (equation \ref{eqn:reflective_scoring}).

Specifically, \method is based on the notion that generated contexts can capture the main semantic aspects of a source text. To explicitly test this, we measure how well the reflective scoring function of equation \ref{eqn:reflective_scoring} captures semantic equivalence on the WMT18 metrics task \citep{ma2018results}. We specifically test on the segment-level evaluation, where the task is to rate the semantic equivalence of a number of machine generated translations to a reference human translation. Metrics/scoring functions are evaluated by their rank correlation (using a metric related to Kendall's Tau \cite{kendall1938new}) with human assessment of semantic equivalence. More details are available in the original task description.

In table \ref{tab:wmt18}, we present results on 3 language pairs: Chinese, German, and Estonian to English. Generally, higher correlation indicates a closer match with human judgement on which translations are closest in meaning to a reference translation. Metrics not significantly outperformed are bolded. We only include to-English translations as the language models used for \method are English. 

To apply equation \ref{eqn:reflective_scoring}, we generate contexts for the reference, and test how well each generated translation fits the reference-contexts. Specifically, we follow equation \ref{eqn:reflective_scoring}, taking the reference sentence as $s_{src}$ and measure the contextual similarity with each translation (taking translation to be $s$ in each case). 

The high performance of our Reflective Scoring function (one of only 2 scoring functions bolded for all 3 pairs) indicates high agreement with human judges semantic equivalence between translations and reference. We are not aiming for state-of-the-art here, but rather to validate that generated contexts can represent full sentences well. This claim seems to be supported. 

\input{figs/wmt18}

\subsection{Ablations}

We include ablation studies for both the number of contexts generated $n_c$ (table \ref{tab:quora_ablation_n_C}) and whether weights $w$ are learned or set to be uniform (table \ref{tab:quora_ablation_weights}). In both cases, besides ablated aspects the experiment is conducted as in \S\ref{sec_experiments}.

For our ablation of generated contexts $n_c$ we investigate all 3 levels of novelty across 4 values of $n_c$: 6, 20, 40, 80, in table \ref{tab:quora_ablation_n_C}. We observe a broad trend of improving metrics with larger values of $n_c$, with the setting used in practice ($n_c = 80$) being a clear winner. An interesting aspect of this is that lower $n_c$ seems to force higher novelty. For $n_c = 20$, the lowest novelty achieved was 37.4, while this was 56.3 for $n_c = 6$. One potential cause for this is that the method cannot sufficiently capture the content of the source sentence with low $n_c$. With $n_c=80$, \method can effectively rewrite the input if it's allowed to, but this doesn't seem to be true for lower $n_c$.

In ablating weight learning, we consider 2 possible values of $n_c$: 6 and 10. This is because, without weight learning we must set $k_c = n_c$ as we cannot do weight pruning if weights are uniform. For $k_c$ much higher than this, the ensemble of equation \ref{eqn:rd} becomes prohibitively expensive to calculate. For both values of $n_c$ we found RD with weight learning outperforms uniform weights on the SARI metric. Interestingly, we found uniform weights resulted in a \textit{higher} novelty at $n_c = 6$ but a \textit{lower} novelty at $n_c = 10$, with a gap of almost 50. With weight learning, this gap is only about 8, indicating less variability in behavior when weight learning is used. 

\input{figs/quora_ablaction_n_c}
\input{figs/quora_ablation_weight_learning}

\section{Further Generations}
\label{app:examples}
See Figures~\ref{fig_quora_examples},\ref{fig_anlg_examples} for outputs of \method and baselines. We also include many generations of \method on the Quora paraphrasing dataset in table \ref{tab:RD_many_Gen}

\input{figs/RD_many_generations}
\input{figs/appendix_gen_anlg_0}
\input{figs/appendix_gen_anlg_1}
\input{figs/appendix_gen_anlg_2}

\input{figs/appendix_gen_paraphrase_0}
\input{figs/appendix_gen_paraphrase_1}

\input{figs/anlg_template}
\input{figs/paraphrasing_template}

%% file: figs/algorithm_lr.tex
\begin{algorithm}[!t]
\small
\centering
\caption{\scriptsize Learn \decoder (left-to-right)}
\label{alg_lr}
\begin{algorithmic}[1]
\REQUIRE Left to right language model \lrLM \\
\hspace{1.2em} Right to left language model \rlLM \\
 \hspace{1.2em} Source text: $s_{src}$
$\ $
\SetAlgoLined
\STATE  Sample contexts, $c_1...c_{n_{c}} \sim \text{\rlLM}(c|s_{src})$\\\
\STATE Initialize parameters $\mathbf{w} = w_1...w_{n_{c}}$ s.t. $\sum w_i = 1, w_i\geq0$\\\
\STATE learn $\mathbf{w}$ = $\arg\max_{\mathbf{w}} \text{\lrRD}(s_{src})$ \\\qquad under $\sum w_i = 1, w_i\geq0$\\\
\ENSURE $\text{\lrRD}$
\end{algorithmic}
\end{algorithm}

%% file: figs/model_params.tex
\begin{table}[t]
\small
    \centering
    \begin{tabular}{l|ccccccc}
     \toprule
      & $len_{s}$ & $len_{c}$ & $n_{s}$ & $n_{c}$ & $h$ & $p_{c}$ & $k_c$ \\
       \midrule
      Pphrase & $inp + 5$& 50 & 30 & 80 & 4. & 0.7 & 6\\
      $\alpha$NLG& 20 & 50 & 20 & 50 & 6. & 0.9 & 6\\ 
    \bottomrule
    \end{tabular}
    \caption{Most parameters are explained in  \S\ref{subsec:implementation}. $h$ is entropy for calibration in \S\ref{app:entropy_calibtration} } %
    \label{tab:params}
\end{table}

%% file: figs/app_paraphrasing_examples.tex
\begin{figure*}[t]
    \centering
    \includegraphics[width=0.9\linewidth]{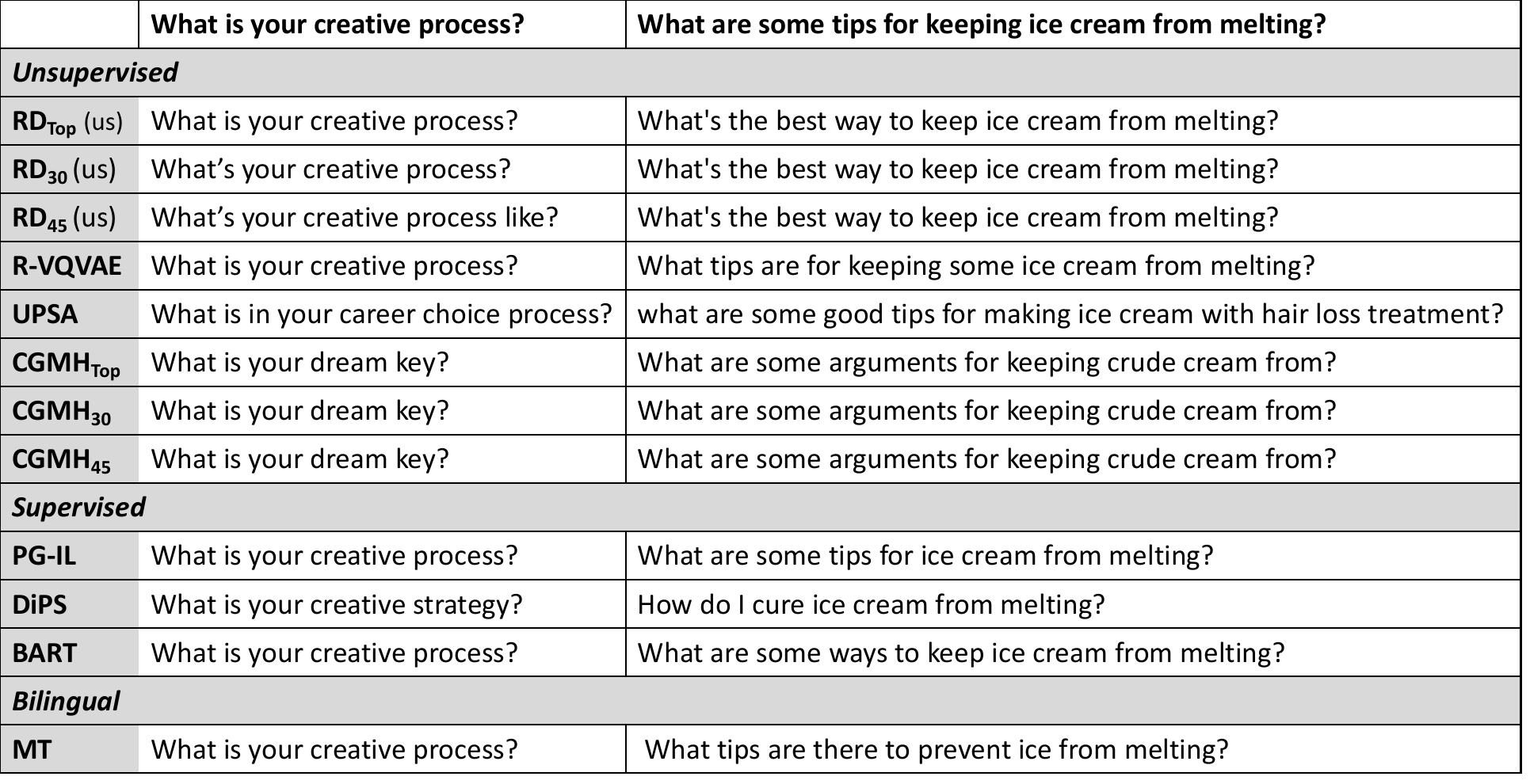}
    \caption{
    Example generations of baselines on Quora paraphrasing dataset (\S\ref{sec_experiments}).
    }
    \label{fig_quora_examples}
\end{figure*} 

%% file: figs/app_anlg_examples.tex
\begin{figure*}[t]
    \centering
    \includegraphics[width=0.95\linewidth]{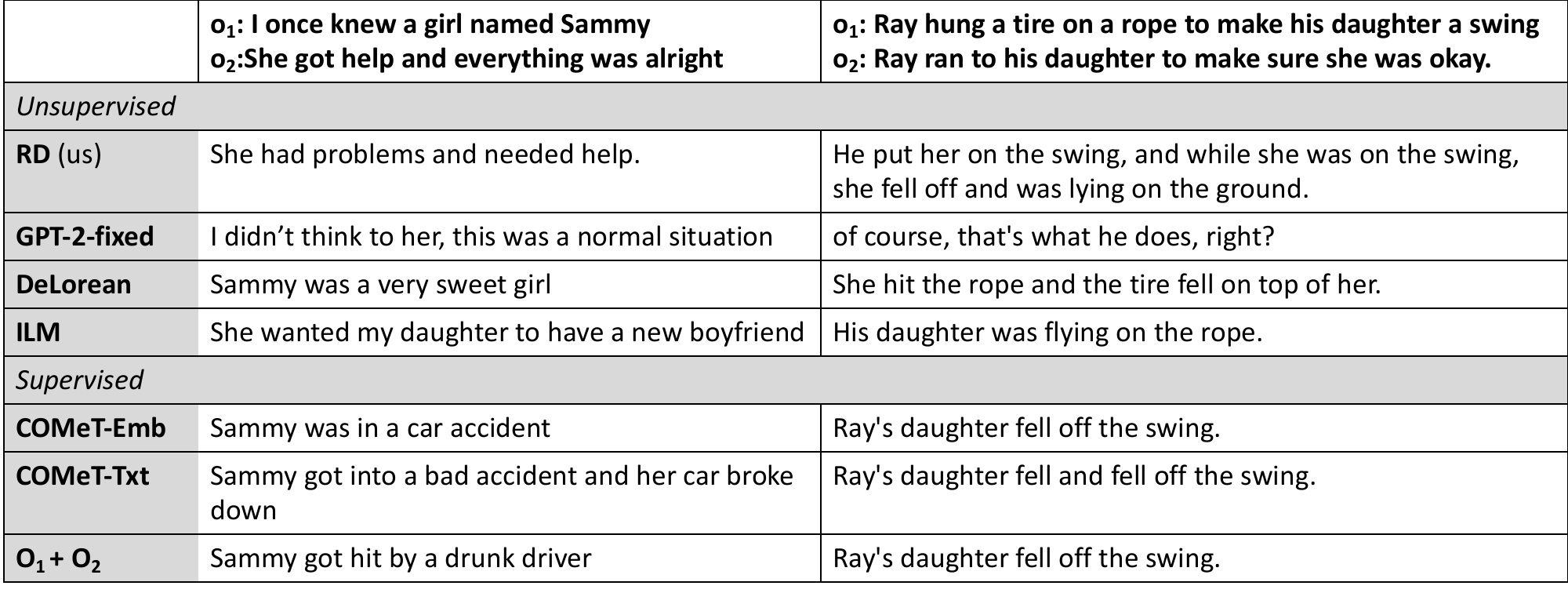}
    \caption{
    Example generations of baselines on $\alpha$NLG dataset (\S\ref{sec_experiments}). Models attempt to fill in the blank between $o_1,o_2$ to explain them both.
    }
    \label{fig_anlg_examples}
\end{figure*} 

%% file: figs/quora_results_aux_metrics.tex
\begin{table*}[t]
\small
    \centering
    \begin{tabular}{ll|ccccc|c}
     \toprule
 & Method & R-1$\uparrow$ & R-2$\uparrow$ & BLEURT$\uparrow$ & BERTScore$\uparrow$ & iBLEU$\uparrow$ &  $Novelty\uparrow$ \\ 
 \midrule
 \textit{Human} & Source & 70.1 & 47.0 & 19.9 & 95.2 & 40.4 & 0.0 \\
   & Reference & 100.0 & 100.0 & 99.3 & 100.0 & 84.4 & 43.9 \\ 
 \midrule
\textit{Supervised} & PG-IL & 66.6 & 44.0 & 11.1 & 94.7 & 36.7 & 24.4 \\
   & DiPS & 56.7 & 33.7 & -29.5 & 92.7 & 31.8 & 48.5\\ 
   & BART & 63.6 & 41.6 & 9.6 & 94.4 & 33.8 & 35.2 \\ 
 \midrule
   \textit{Supervised (Bilingual)} & MT & 64.7 & 39.8 & 16.7 & 94.8 & 35.9 & 26.8\\
   \midrule
  \textit{Unsupervised} & R-VQVAE & 68.2 & 32.0 & -7.6 & 93.2 & 31.9 & 26.2\\ 
   & CGMH$_{Top}$& 55.6 & 29.6 & -53.6 & 92.1 & 30.6 & 27.6\\ 
   & CGMH$_{30}$ & 54.5 & 28.3 & -58.9 & 91.9 & 29.8 & 29.7\\ 
   & CGMH$_{45}$ & 48.5 & 22.1 & -80.9 & 90.7 & 24.9 & 44.5\\ 
   & UPSA & 56.2 & 30.4 & -44.5 & 90.7 & 27.3& 44.4 \\
  & RD$_{Top}$ (Us) & 65.8 & 42.3 & 15.3 & 94.8 & 37.0 & 20.8\\
   & RD$_{30}$ (Us) & 62.1 & 38.0 & 7.7 & 94.2 & 35.1 & 30.0\\
   & RD$_{45}$ (Us) & 56.8 & 31.1 & -1.9 & 93.5 & 30.4 & 45.0\\
    \bottomrule
    \end{tabular}
    \caption{Model performance on the Quora test split. Included here are extra metrics beyond what is in the main paper. R-1 and R-2 refer to ROUGE-1 and ROUGE-2.}
    
    \label{tab:quoratest_aux}
\end{table*}

%% file: figs/combined_human.tex
\begin{table*}[t]
\small
    \centering
    \begin{tabular}{l|ccc|c||ccc|c}
     \toprule
       & \multicolumn{4}{c}{Quora} & \multicolumn{4}{c}{Twitter}\\
 Method & Fluency & Consistency & Novelty &  Overall (\%) & Fluency & Consistency & Novelty &  Overall (\%) \\ 
 \midrule
 \midrule
   Reference & 98.7 & 78.3 & 94.0 & 71.7 & 91.7 & 58.7 & 95.3 & 51.3 \\ 
 \midrule
 \midrule
PG-IL & 95.9 & 79.9 & 51.0 & 29.4 & - & - & - & - \\
   DiPS &  85.6 & 45.1 & 93.3 & 36.6 & - & - & - & - \\ 
   BART & 97.2 & 77.6 & 68.8 & 46.1 & - & - & - & - \\ 
    \midrule
 \midrule
   MT & 98.7 & 88.7 & 71.2 & 59.3 & 99.0 & 90.0 & 80.9 & 70.9 \\
   \midrule
   \midrule
   R-VQVAE & 84.2 & 76.3 & 60.3 & 33.5 & 65.3 & 44.0 & 94.3 & 32.3 \\ 
   CGMH$_{Top}$ & 79.4 & 43.1 & 85.6 & 27.0 & 71.9 & 48.8 & 82.6 & 27.8 \\ 
   CGMH$_{30}$ &  78.8 & 37.9 & 96.4 & 31.5 & 67.2 & 35.8 & 92.0 & 25.1 \\ 
   CGMH$_{45}$ & 71.6 & 19.9 & 98.5 & 15.8 & 51.5 & 20.9 & 96.3 & 13.5 \\ 
   UPSA	& 84.4 & 46.7 & 91.6 & 37.8 & - & - & - & - \\
   \midrule
   RD$_{Top}$ (Us) & 98.0 & 84.6 & 43.5 & 27.5 & 98.7 & 70.9 & 76.3 & 46.5\\
   RD$_{30}$ (Us) & 98.7 & 75.3 & 88.2 & 63.2 & 98.7 & 70.9 & 76.3 & 46.5 \\
   RD$_{45}$ (Us) & 97.5 & 67.3 & 95.3 & 62.1 & 98.0 & 64.5 & 92.6 & 56.9 \\

    \bottomrule
    \end{tabular}
    \caption{Human evaluation results on both datasets for tested models. See \S\ref{app:human_metrics}.}%
    \label{tab:combined_human}
\end{table*}

%% file: figs/quora_human.tex
\begin{table*}[t]
\small
    \centering
    \begin{tabular}{l|ccc|c}
     \toprule
       & \multicolumn{4}{c}{Human Quality$\uparrow$}\\
 Method & Fluency & Consistency & Novelty &  Overall (\%) \\ 
 \midrule
  \rowcolor[gray]{0.95} \multicolumn{5}{l}{\textit{Human}} \\
 \midrule
   Reference & 98.7 & 78.3 & 94.0 & 71.7 \\ 
 \midrule
 \rowcolor[gray]{0.95} \multicolumn{5}{l}{\textit{Supervised}} \\
 \midrule
PG-IL & 95.9 & 79.9 & 51.0 & 29.4  \\
   DiPS &  85.6 & 45.1 & 93.3 & 36.6 \\ 
   BART & 97.2 & 77.6 & 68.8 & 46.1  \\ 
    \midrule
 \rowcolor[gray]{0.95} \multicolumn{5}{l}{\textit{Bilingual}} \\
 \midrule
   MT & 98.7 & 88.7 & 71.2 & 59.3 \\
   \midrule
   \rowcolor[gray]{0.95} \multicolumn{5}{l}{\textit{Unsupervised}} \\
   \midrule
   R-VQVAE & 84.2 & 76.3 & 60.3 & 33.5 \\ 
   CGMH$_{Top}$ & 79.4 & 43.1 & 85.6 & 27.0 \\ 
   CGMH$_{30}$ &  78.8 & 37.9 & 96.4 & 31.5 \\ 
   CGMH$_{45}$ & 71.6 & 19.9 & 98.5 & 15.8 \\ 
   UPSA	& 84.4 & 46.7 & 91.6 & 37.8 \\
   \midrule
   RD$_{Top}$ (Us) & 98.0 & 84.6 & 43.5 & 27.5\\
   RD$_{30}$ (Us) & 98.7 & 75.3 & 88.2 & 63.2 \\
   RD$_{45}$ (Us) & 97.5 & 67.3 & 95.3 & 62.1 \\

    \bottomrule
    \end{tabular}
    \caption{Model performance on the Quora test split, by human evaluation. Overall is calculated as the percentage of generations that meet the basic criteria of a paraphrase: fluent  (the paraphrase can be understood), consistent with the source (the paraphrase shows at most \textbf{minor differences} in meaning from the source) and giving a novel phrasing (paraphrase shows at least \textbf{minor difference} in word choice). The first 3 columns indicate percentage of generations that meet the given criterion. Note, the first 3 rows (fluency, consistency, and novelty) are all required to for our notion of a good paraphrase, and each can be trivially maximized on its own.}%
    \label{tab:quora_human}
\end{table*}

%% file: figs/twitter_results.tex
\begin{table*}[t]
\small
    \centering
    \begin{tabular}{ll|cccc|c}
     \toprule
 & Method & SARI$\uparrow$ & BLEU$\uparrow$ & METEOR$\uparrow$ & TER$_{P}$$\downarrow$ &  $Novelty\uparrow$ \\ 
 \midrule
 \textit{Human} & Source & 13.6 & 36.7 & 25.0 & 75.0 & 0.0 \\
   & Reference & 90.7 & 100.0 & 100.0 & 0.0 & 63.3 \\ 
 \midrule
   \textit{Supervised (Bilingual)} & MT & 36.1 & 29.4 & 22.1 & 80.0 & 30.4\\
   \midrule
  \textit{Unsupervised} & R-VQVAE & 31.1 & 25.2 & 21.0 & 90.0 & 40.4\\ 
   & CGMH$_{Top}$&  32.7 & 28.2 & 19.8 & 77.0 & 25.5\\
   & CGMH$_{30}$&  33.2 & 26.3 & 18.7 & 78.0 & 30.1\\
   & CGMH$_{45}$&  31.8 & 20.5 & 15.4 & 82.0 & 45.7\\
  & RD$_{Top/30}$ (Us) & 31.4 & 27.2 & 19.9 & 86.0 & 37.0\\
   & RD$_{45}$ (Us) & 36.1 & 25.6 & 19.0 & 88.0 & 45.3\\
    \bottomrule
    \end{tabular}
    \caption{Model performance on the Twitter URL test split. Note: Diversity of RD$_{Top}$ is over 30 and so this model is equivalent to RD$_{30}$ here.}
    
    \label{tab:twitter_results}
\end{table*}

%% file: figs/twitter_human.tex
\begin{table*}[t]
\small
    \centering
    \begin{tabular}{l|ccc|c}
     \toprule
       & \multicolumn{4}{c}{Human Quality$\uparrow$}\\
 Method & Fluency & Consistency & Novelty &  Overall \\ 
 \midrule
  \rowcolor[gray]{0.95} \multicolumn{5}{l}{\textit{Human}} \\
 \midrule
   Reference & 91.7 & 58.7 & 95.3 & 51.3 \\ 
    \midrule
 \rowcolor[gray]{0.95} \multicolumn{5}{l}{\textit{Bilingual}} \\
 \midrule
   MT  & 99.0 & 90.0 & 80.9 & 70.9 \\
   \midrule
   \rowcolor[gray]{0.95} \multicolumn{5}{l}{\textit{Unsupervised}} \\
   \midrule
   R-VQVAE & 65.3 & 44.0 & 94.3 & 32.3 \\ 
   CGMH$_{Top}$ & 71.9 & 48.8 & 82.6 & 27.8 \\ 
   CGMH$_{30}$ &  67.2 & 35.8 & 92.0 & 25.1 \\ 
   CGMH$_{45}$ & 51.5 & 20.9 & 96.3 & 13.5 \\ 
   \midrule
   RD$_{Top/30}$ (Us) & 98.7 & 70.9 & 76.3 & 46.5 \\
   RD$_{45}$ (Us) & 98.0 & 64.5 & 92.6 & 56.9 \\

    \bottomrule
    \end{tabular}
    \caption{Model performance on the Twitter test split, by human evaluation. Overall is calculated as the percentage of generations that meet the basic criteria of a paraphrase: fluent  (the paraphrase can be understood), consistent with the source (the paraphrase shows at most \textbf{minor differences} in meaning from the source) and giving a novel phrasing (paraphrase shows at least \textbf{minor difference} in word choice). The first 3 columns indicate percentage of generations that meet the given criterion. Note, the first 3 rows (fluency, consistency, and novelty) are all required to for our notion of a good paraphrase, and each can be trivially maximized on its own.}%
    \label{tab:twitter_human}
\end{table*}

%% file: figs/wmt18.tex
\begin{table*}[t]
\small
    \centering
    \begin{tabular}{l|ccc}
     \toprule
Metric & cs$\xrightarrow{}$en & de$\xrightarrow{}$en & et$\xrightarrow{}$en \\ 

 \midrule
Reflective Score & \textbf{0.357}  & \textbf{0.490} & \textbf{0.364} \\
chrF+ & 0.288 & 0.479 & 0.332 \\
sentBLEU  & 0.233 & 0.415 & 0.285 \\
CharacTER &  0.256 & 0.450 & 0.286 \\
BEER & 0.295 & 0.481 & 0.341 \\
ITER &  0.198 & 0.396 & 0.235 \\
RUSE & \textbf{0.347} & \textbf{0.498} & \textbf{0.368} \\
chrF & 0.288 & 0.479 & 0.328 \\
meteor++ & 0.270 & 0.457 & 0.329 \\
YiSi-1 & \textbf{0.319} & 0.488 & 0.351 \\
YiSi-0 & 0.301 & 0.474 & 0.330 \\
BLEND & \textbf{0.322} & 0.492 & 0.354 \\ 
YiSi-1$_srl$ & \textbf{0.317} & \textbf{0.483} & 0.345 \\
UHH$_TSKM$ & 0.274 & 0.436 & 0.300 \\
    \bottomrule
    \end{tabular}
    \caption{Correlation with human judgement on the WMT18 metric task, for 3 language pairs (Chinese, German, and Estonian to English). Correlations of metrics not significantly outperformed by any other for that language pair are highlighted in bold. We note that Reflective Score is among only 2 methods bolded over all language-pairs tested.}
    
    \label{tab:wmt18}
\end{table*}

%% file: figs/quora_ablaction_n_c.tex
\begin{table*}[t]
\small
    \centering
    \begin{tabular}{ll|cccc|c}
     \toprule
Method & $n_c$ & SARI$\uparrow$ & BLEU$\uparrow$ & METEOR$\uparrow$ & TER$_{P}$$\downarrow$ &  $Novelty\uparrow$ \\ 

 \midrule
RD$_{Top}$ & 80 & 29.0 & 49.9 & 33.9 & 52.0  & 20.8\\
& 40 & 31.3 & 46.6 & 31.8 & 56.0 & 28.4 \\
& 20 & 34.1 & 43.0 & 29.8 & 61.0 & 37.4 \\
& 6 & 33.3 &	33.2 & 23.8 & 71.0 & 56.3\\

 \midrule

RD$_{30}$ &  80 & 40.0 & 46.8 & 32.2 & 57.0 & 30.0\\
& 40 & 39.4 & 44.2 & 30.6 & 59.0 & 35.6 \\
& 20 & 34.1 & 43.0 & 29.8 & 61.0 & 37.4 \\
& 6 & 33.3 &	33.2 & 23.8 & 71.0 & 56.3\\
   \midrule
  RD$_{45}$ & 80 & 38.6 & 39.9 & 28.9 & 65.0 &  45.0\\
  & 40 & 38.3 & 39.3 & 28.0 & 65.0 & 46.1 \\
  & 20 & 38.5 & 39.6 & 28.0 & 65.0 & 45.8 \\
  & 6 & 33.3 &	33.2 & 23.8 & 71.0 & 56.3\\
    \bottomrule
    \end{tabular}
    \caption{Ablation of number of contexts generated $n_c$, holding weight pruning parameter constant at $k_c = 6$. For some $n_c$ (e.g. $n_c = 6$), the \textit{Top} setting achieves high enough novelty for both cutoffs (30 and 45). In these cases, $RD_{Top}$ is repeated for $RD_{30}$ and $RD_{45}$}
    
    \label{tab:quora_ablation_n_C}
\end{table*}

%% file: figs/quora_ablation_weight_learning.tex
\begin{table*}[t]
\small
    \centering
    \begin{tabular}{l|cccc|c}
     \toprule
Method &  SARI$\uparrow$ & BLEU$\uparrow$ & METEOR$\uparrow$ & TER$_{P}$$\downarrow$ &  $Novelty\uparrow$ \\ 

 \midrule
RD$_{Top}$ ($n_c = 6$ ) & 33.3 &	33.2 & 23.8 & 71.0 & 56.3\\
\textit{- weight learning} & 29.6 & 20.1 & 16.8 & 89.0 & 76.7\\

 \midrule
RD$_{Top}$ ($n_c = 10$ ) & 34.2 & 37.3 & 26.4 & 65.0 & 48.2\\
\textit{- weight learning} & 31.3 & 46.6 & 31.8 & 56.0 & 28.4\\

    \bottomrule
    \end{tabular}
    \caption{Ablation of whether weights are learned or taken to be uniform. For learned weights, we set $k_c = n_c$ indicating no weight pruning, so that the final number of contexts used is the same with and without learning. }
    
    \label{tab:quora_ablation_weights}
\end{table*}

%% file: figs/RD_many_generations.tex
\begin{table*}[t]
\small
    \centering
\begin{tabular}{l|c}
\midrule
\rowcolor[gray]{0.95}Input&what are the best books to expand imagination ?\\
\midrule
RD$_{Top}$& what are the best books to expand imagination ?\\
RD$_{30}$& what are the best books to expand \textbf{our} imagination ?\\
RD$_{45}$& what books \textbf{should you read} to expand \textbf{your} imagination ?\\
\midrule
\rowcolor[gray]{0.95}Input&how safe are ready to eat products ?\\
\midrule
RD$_{Top}$& how safe are ready to eat \textbf{foods and} products ?\\
RD$_{30}$& how safe are ready to eat \textbf{foods and} products ?\\
RD$_{45}$& how safe are foods that \underline{are ready to eat} ?\\
\midrule
\rowcolor[gray]{0.95}Input&what can be the future of kashmir ?\\
\midrule
RD$_{Top}$& what \textbf{is} the future of kashmir ?\\
RD$_{30}$& what \textbf{is} the future of kashmir ?\\
RD$_{45}$& what \textbf{will} be the \textbf{fate} of kashmir ?\\
\midrule
\rowcolor[gray]{0.95}Input&does eating eggs also cause bird flu ?\\
\midrule
RD$_{Top}$& \textbf{is it true that} eating eggs \textbf{can cause} bird flu ?\\
RD$_{30}$& \textbf{is it true that} eating eggs \textbf{can cause} bird flu ?\\
RD$_{45}$& \textbf{is it true that} eating eggs \textbf{can cause} bird flu ?\\
\midrule
\rowcolor[gray]{0.95}Input&how do you open odt files on word ?\\
\midrule
RD$_{Top}$& how do you open odt files on word ?\\
RD$_{30}$& how do you open odt files on \textbf{microsoft} word ?\\
RD$_{45}$& how do you open an odt file on word ?\\
\midrule
\rowcolor[gray]{0.95}Input&if someone rejects your facebook friend request can you send it again ?\\
\midrule
RD$_{Top}$& if someone rejects your facebook friend request can you send it again ?\\
RD$_{30}$& if someone rejects your facebook friend request can you \underline{message} them again ?\\
RD$_{45}$& \underline{what if someone accepts} your facebook friend request ?\\
\midrule
\rowcolor[gray]{0.95}Input&when will science prove that god exists ?\\
\midrule
RD$_{Top}$& when will science prove that god exists ?\\
RD$_{30}$& when will science prove \textbf{the existence of} god ?\\
RD$_{45}$& when will science prove \textbf{the existence of} god ?\\
\midrule
\rowcolor[gray]{0.95}Input&is it good to use hair dryers ?\\
\midrule
RD$_{Top}$& is it good to use hair dryers ?\\
RD$_{30}$& is it good to use \textbf{a} hair dryer ?\\
RD$_{45}$& \underline{do you know that it 's good} to use hair dryers ?\\
\midrule
\rowcolor[gray]{0.95}Input&is demonetizing notes of 500 and 1000 a permanent solution to curb black money ?\\
\midrule
RD$_{Top}$& is \underline{it} a permanent solution to curb black money ?\\
RD$_{30}$& is \underline{it} a permanent solution to curb black money ?\\
RD$_{45}$& \textbf{do you think that} demonetizing rs 500 and rs 1000 notes \textbf{is going to solve the problem of} black money ?\\
\midrule
\rowcolor[gray]{0.95}Input&what 's the best way to ask out a girl at my school ?\\
\midrule
RD$_{Top}$& what 's the best way to ask out a girl \textbf{in} my school ?\\
RD$_{30}$& what 's the best way to ask out a girl \textbf{in} my school ?\\
RD$_{45}$& \textbf{how do i ask a girl out} at school ?\\
\midrule
\rowcolor[gray]{0.95}Input&what was meant by the final scene in 2001 : a space odyssey ?\\
\midrule
RD$_{Top}$& what was meant by the final scene in 2001 : a space odyssey ?\\
RD$_{30}$& what did he mean by the final scene in 2001 : a space odyssey ?\\
RD$_{45}$& what did \textbf{stanley kubrick} mean by the final scene in \textbf{2001} ?\\
\midrule
\rowcolor[gray]{0.95}Input&how can i help my dog get rid of hiccups ?\\
\midrule
RD$_{Top}$& how can i help my dog get rid of hiccups ?\\
RD$_{30}$& how can i help my dog get rid of \textbf{the} hiccups ?\\
RD$_{45}$& can i help my dog \textbf{with hiccups} ?\\

\bottomrule
\end{tabular}

    \caption{Generations from the 3 \method variants on the examples from the Quora test split. Places where a novel phrasing is introduced are \textbf{bolded} and mistakes are \underline{underlined}.}  %
    \label{tab:RD_many_Gen}

\end{table*}

%% file: figs/appendix_gen_anlg_0.tex
\begin{table*}[t]
\small
    \centering
\begin{tabular}{l|c}
\toprule

\rowcolor[gray]{0.95} \multicolumn{2}{c}{ Sam didn't sleep well last night. \_\_\_ Sam felt embarrassed and was planning on going to bed early that day.
}\\
\midrule
\textbf{Human} & Sam treated people poorly due to lack of sleep. \\
\midrule
\rowcolor[gray]{0.95} \multicolumn{2}{l}{Unsupervised}\\
\midrule
\textbf{\method{}} & He was so tired that he couldn't even get out of bed. \\
\midrule
\textbf{GPT2-fixed} & he was really tired from finishing his book \\
\midrule
\rowcolor[gray]{0.95} \multicolumn{2}{l}{Supervised}\\
\midrule
\textbf{COMeT-Emb+GPT2} & Sam was still tired when she woke up. \\
\midrule
\textbf{COMeT-Txt+GPT2} & Sam was too tired to finish his homework. \\
\midrule
\textbf{$O_1$-$O_2$-Only} & Sam was very tired after dinner. \\
\bottomrule
\end{tabular}

    \caption{Further generations for $\alpha$NLG from all tested systems}  %
    \label{tab:anlg_gn_0}

\end{table*}

%


%% file: figs/appendix_gen_anlg_1.tex
\begin{table*}[t]
\small
    \centering
\begin{tabular}{l|c}
\toprule

\rowcolor[gray]{0.95} \multicolumn{2}{c}{ Cindy got out of bed. \_\_\_ She also ordered an iPhone.}\\
\midrule
\textbf{Human} & She ordered a new iPad.  \\
\midrule
\rowcolor[gray]{0.95} \multicolumn{2}{l}{Unsupervised}\\
\midrule
\textbf{\method{}} & On her way out, she ordered an iPad, one that was small enough to fit in her apartment. \\
\midrule
\textbf{GPT2-fixed} & I'm sure, given all the likely possibilities of electrical shocks and heart attacks it's a good thing \\
\midrule
\rowcolor[gray]{0.95} \multicolumn{2}{l}{Supervised}\\
\midrule
\textbf{COMeT-Emb+GPT2} & Cindy went on Facebook. \\
\midrule
\textbf{COMeT-Txt+GPT2} & Cindy liked the new slim iPhone. \\
\midrule
\textbf{$O_1$-$O_2$-Only} &  Cindy saw she didn't have enough money for clothes. \\
\bottomrule
\end{tabular}

    \caption{Further generations for $\alpha$NLG from all tested systems}  %
    \label{tab:anlg_gn_1}

\end{table*}

%% file: figs/appendix_gen_anlg_2.tex
\begin{table*}[t]
\small
    \centering
\begin{tabular}{l|c}
\toprule

\rowcolor[gray]{0.95} \multicolumn{2}{c}{  I once knew a girl named Sammy. \_\_\_ She got help and everything was alright.}\\
\midrule
\textbf{Human} &  She suffered from depression.  \\
\midrule
\rowcolor[gray]{0.95} \multicolumn{2}{l}{Unsupervised}\\
\midrule
\textbf{\method{}} & She had problems and she needed help. \\
\midrule
\textbf{GPT2-fixed} & I didn't think to her, this was a normal situation \\
\midrule
\rowcolor[gray]{0.95} \multicolumn{2}{l}{Supervised}\\
\midrule
\textbf{COMeT-Emb+GPT2} & Sammy was in a car accident. \\
\midrule
\textbf{COMeT-Txt+GPT2} & Sammy got into a bad accident and her car broke down. \\
\midrule
\textbf{$O_1$-$O_2$-Only} &  Sammy got hit by a drunk driver. \\
\bottomrule
\end{tabular}

    \caption{Further generations for $\alpha$NLG from all tested systems}  %
    \label{tab:anlg_gn_2}

\end{table*}

%% file: figs/appendix_gen_paraphrase_0.tex
\begin{table*}[t]
\small
    \centering
\begin{tabular}{l|c}
\toprule

\rowcolor[gray]{0.95} \multicolumn{2}{c}{Can you trust the information on Quora?}\\
\midrule
\textbf{Human} & Do you trust Quora? \\
\midrule
\rowcolor[gray]{0.95} \multicolumn{2}{l}{Unsupervised}\\
\midrule
\textbf{RefDec-Top (Us)} & Can you trust the information on Quora?
 \\
\midrule
\textbf{RefDec-70 (Us)} & Can I trust the information on Quora? \\
\midrule
\textbf{RefDec-55 (Us)} & When can I trust information on Quora ? \\
\midrule
\midrule
\textbf{R-VQVAE} & Can you trust the information on Quora?\\
\midrule
\textbf{CGMH-Top} & Can you answer the information on Quora? \\
\midrule
\textbf{CGMH-70} & Can you answer the information on Quora? \\
\midrule
\textbf{CGMH-55} & Can you answer more topics on Quora? \\
\midrule
\rowcolor[gray]{0.95} \multicolumn{2}{l}{Supervised}\\
\midrule
\textbf{PG-IL} & Can you trust the information on Quora?
 \\
\midrule
\textbf{DiPS} & Can we trust our questions in Quora? \\
\midrule
\textbf{BART} & Can you trust everything you read on Quora?
 \\
\midrule
\rowcolor[gray]{0.95} \multicolumn{2}{l}{Bilingual} \\
\midrule
\textbf{MT} & Can you trust the information on Quora? \\
\bottomrule
\end{tabular}

    \caption{Further generations for paraphrasing from all tested systems}  %
    \label{tab:paraphrase_gen_0}

\end{table*}

%% file: figs/appendix_gen_paraphrase_1.tex
\begin{table*}[t]
\small
    \centering
\begin{tabular}{l|c}
\toprule

\rowcolor[gray]{0.95} \multicolumn{2}{c}{ What is your creative process? }\\
\midrule
\textbf{Human} & What's your creative process? \\
\midrule
\rowcolor[gray]{0.95} \multicolumn{2}{l}{Unsupervised}\\
\midrule
\textbf{RefDec-Top (Us)} & What is your creative process?  \\
\midrule
\textbf{RefDec-70 (Us)} & What's your creative process?  \\
\midrule
\textbf{RefDec-55 (Us)} & What's your creative process like? \\
\midrule
\midrule
\textbf{R-VQVAE} & What is your creative process? \\
\midrule
\textbf{CGMH-Top} &  What is your dream key? \\
\midrule
\textbf{CGMH-70} &  What is your dream key? \\
\midrule
\textbf{CGMH-55} & What is your dream key? \\
\midrule
\rowcolor[gray]{0.95} \multicolumn{2}{l}{Supervised} \\
\midrule
\textbf{PG-IL} & What is your creative process? \\
\midrule
\textbf{DiPS} & What is your creative strategy? \\
\midrule
\textbf{BART} & What is your creative process? \\
\midrule
\rowcolor[gray]{0.95} \multicolumn{2}{l}{Bilingual} \\
\midrule
\textbf{MT} & What is your creative process? \\
\bottomrule
\end{tabular}

    \caption{Further generations for paraphrasing from all tested systems}  %
    \label{tab:paraphrase_gen_1}

\end{table*}

%% file: figs/anlg_template.tex
\begin{figure*}[t]
    \centering
    \includegraphics[width=0.8\linewidth]{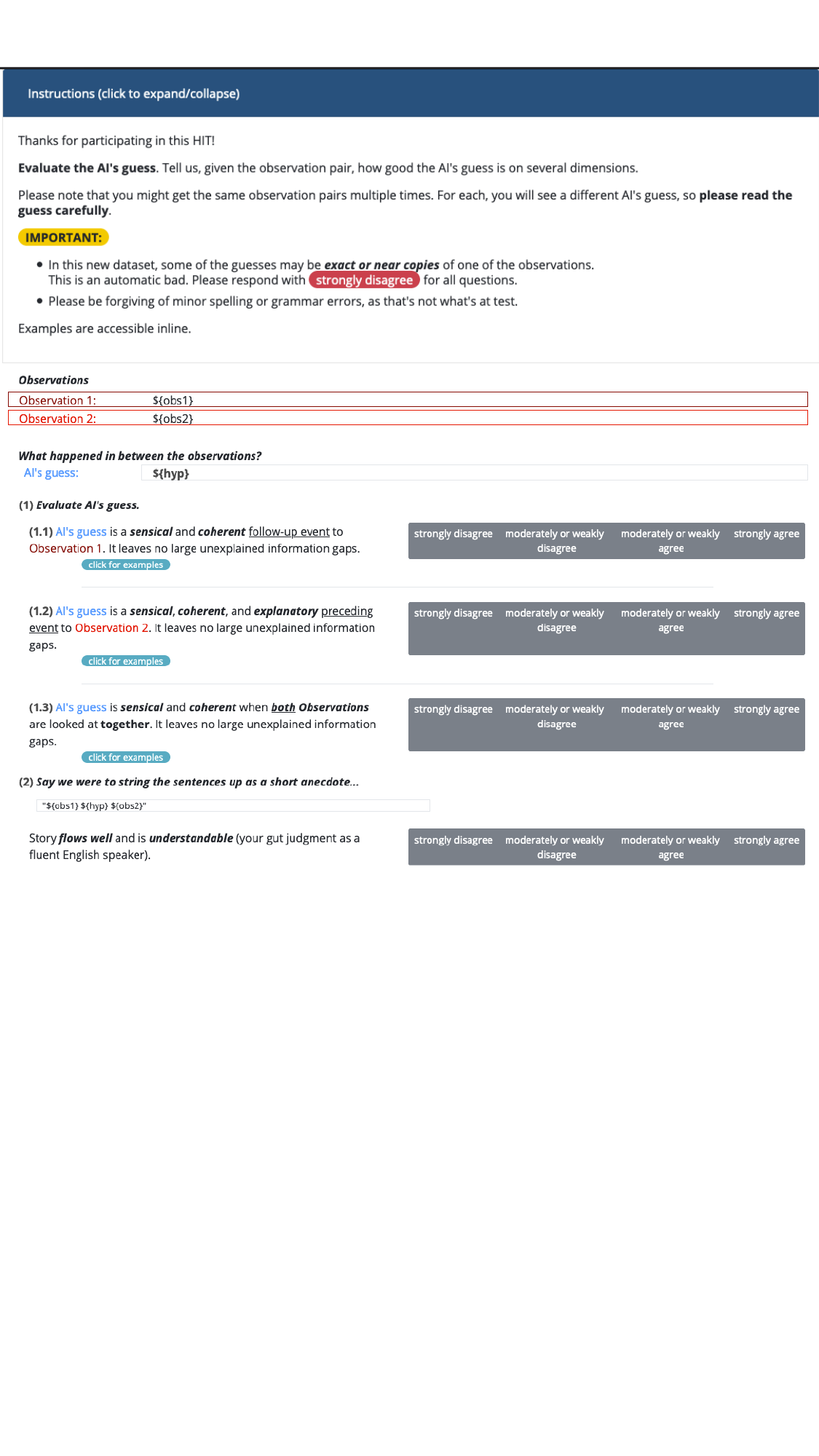}
    \caption{The template used for human evaluation of $\alpha$NLG
    }
    \label{fig:anlg_template}
\end{figure*} 

%% file: figs/paraphrasing_template.tex
\begin{figure*}[t]
    \centering
    \includegraphics[width=0.8\linewidth]{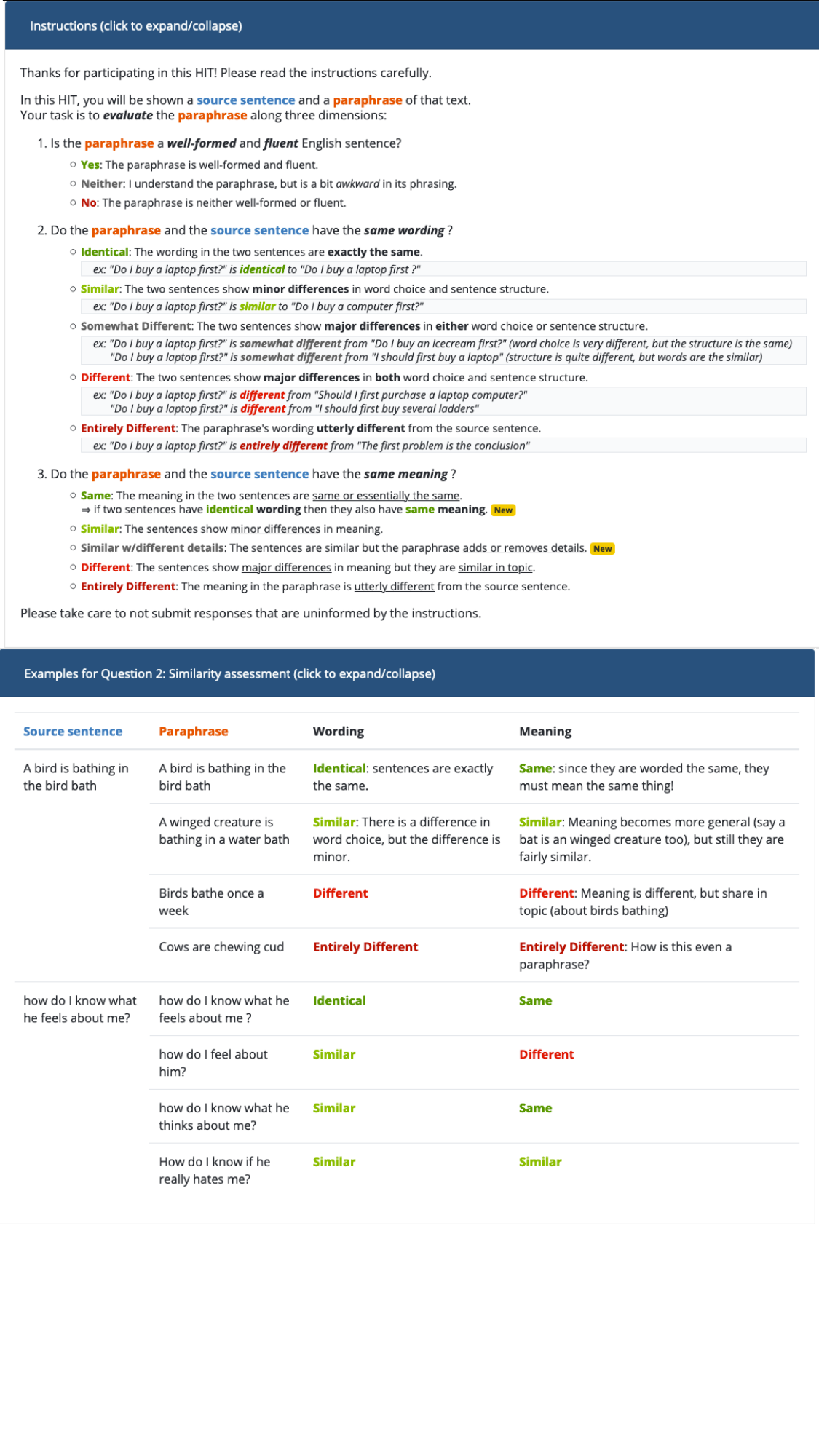}
    \caption{The template used for human evaluation of paraphrasing (part 1 of 2)
    }
    \label{fig:paraphrasing_template}
\end{figure*} 

\begin{figure*}[t]
    \centering
    \includegraphics[width=0.8\linewidth]{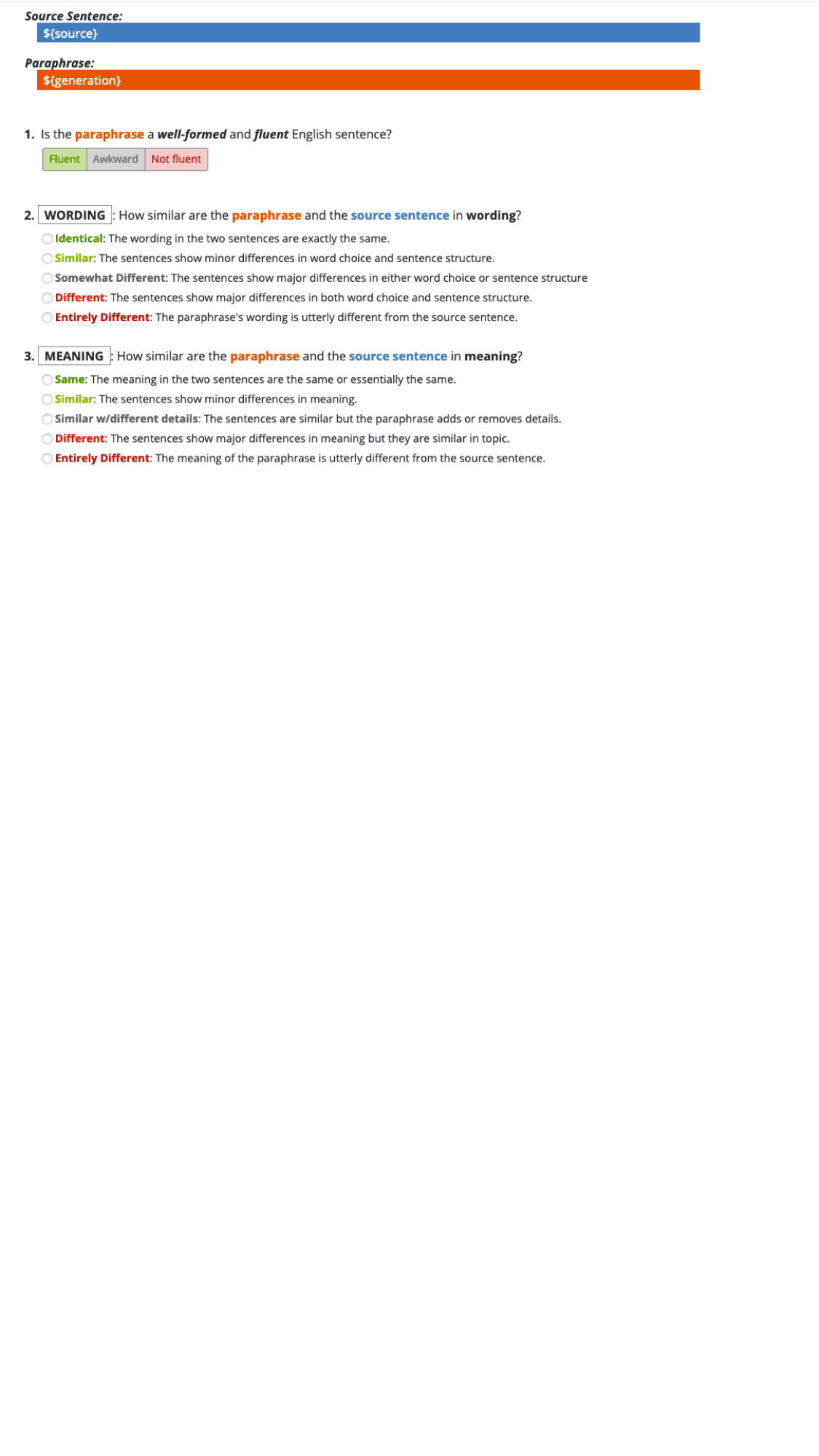}
    \caption{The template used for human evaluation of paraphrasing (part 2 of 2)
    }
    \label{fig:paraphrasing_template_0}
\end{figure*} 